\documentclass{article} 
\usepackage{iclr2024_conference,times}

\usepackage{hyperref}
\usepackage{url}
\usepackage{graphicx}
\usepackage{subcaption} 
\usepackage{multirow} 
\usepackage{listings} 
\usepackage{booktabs}
\usepackage{pifont} 
\usepackage{xcolor}
\usepackage{amsmath}
\newcommand{\cmark}{\ding{51}}%
\newcommand{\xmark}{\ding{55}}%
\newcommand{\cmarkg}{\textcolor{green!70!black}{\cmark}} 
\newcommand{\xmarkr}{\textcolor{red!70!black}{\xmark}}   

\title{TongSIM: A General Platform for Simulating Intelligent Machines}

\iclrfinalcopy
\author{Zhe Sun, Kunlun Wu, Chuanjian Fu, Zeming Song, Langyong Shi, Zihe Xue, Bohan Jing, \\
\textbf{Ying Yang, Xiaomeng Gao, Aijia Li, Tianyu Guo, Huiying Li, Xueyuan Yang, Rongkai Liu,} \\
\textbf{Xinyi He, Yuxi Wang, Yue Li, Mingyuan Liu, Yujie Lu, Hongzhao Xie, Shiyun Zhao, Bo Dai,}\\
\textbf{Wei Wang, Tao Yuan, Song-Chun Zhu, Yujia Peng, Zhenliang Zhang} \\
\textbf{} \\
\centerline{State Key Laboratory of General Artificial Intelligence, BIGAI, Beijing, China} \\
 \\
}

\renewcommand{\headrulewidth}{0pt}

\begin{document}

\maketitle

\begin{abstract}
As artificial intelligence (AI) rapidly advances, especially in multimodal large language models (MLLMs), research focus is shifting from single-modality text processing to the more complex domains of multimodal and embodied AI. Embodied intelligence focuses on training agents within realistic simulated environments, leveraging physical interaction and action feedback rather than conventionally labeled datasets. Yet, most existing simulation platforms remain narrowly designed, each tailored to specific tasks. A versatile, general-purpose training environment that can support everything from low-level embodied navigation to high-level composite activities, such as multi-agent social simulation and human-AI collaboration, remains largely unavailable.
To bridge this gap, we introduce TongSIM, a high-fidelity, general-purpose platform for training and evaluating embodied agents. TongSIM offers practical advantages by providing over 100 diverse, multi-room indoor scenarios as well as an open-ended, interaction-rich outdoor town simulation, ensuring broad applicability across research needs. Its comprehensive evaluation framework and benchmarks enable precise assessment of agent capabilities, such as perception, cognition, decision-making, human-robot cooperation, and spatial and social reasoning. With features like customized scenes, task-adaptive fidelity, diverse agent types, and dynamic environmental simulation, TongSIM delivers flexibility and scalability for researchers, serving as a unified platform that accelerates training, evaluation, and advancement toward general embodied intelligence.
The source code is publicly available at \url{https://github.com/bigai-ai/tongsim}.

\end{abstract}

\section{Introduction}

The emergence of large language models (LLMs) has revolutionized the understanding of artificial intelligence (AI). Researchers quickly uncovered a diverse array of capabilities of LLMs within the text modal, including multi-turn dialogue agents \cite{dulac2021challenges, ahn2022can, sinha2024challenges}, automatic customer services \cite{meincke2025reimagining, shi2024chops}, novel generation and completion \cite{shakeri2021saga, li2024we}, text-based automatic non-player characters (NPCs) \cite{bailis2024werewolf, zhang2025dvm}, role-playing \cite{park2023generative, rao2024collaborative}, and AI-engaged educational systems \cite{bhattacharjee2024understanding, chu2025llm, cordova2025ai}. As the competence of LLMs rapidly increases, expectations for these models have begun to shift from purely text-based applications toward multimodal extensions.

Consequently, a series of research works focused on connecting AI to the real world has emerged \cite{seo2023masked, park2023generative, fang2023generalization, zhang2018iar}. A particularly prominent area within this trend is Embodied AI. Rather than relying on conventional labeled data, researchers in this field propose that training agents in high-fidelity simulated environments and providing them with an embodiment and corresponding action-based feedback can inject new vitality into AI development \cite{roy2021machine}. A sufficiently realistic simulation environment is crucial to this field. This necessity has spurred the development of a variety of simulation platforms, such as Interactive Gibson \cite{xia2020interactive} and MINOS \cite{savva2017minos} that focused on navigation, VRGym \cite{xie2019vrgym} addressing various human interfaces for multimodal interaction, and platforms dedicated to indoor household tasks \cite{shridhar2020alfworld, hong2024multiply, puig2020watch}. While some simulation platforms primarily support only one or a few categories of embodied AI tasks, the scope of supported task types is observably increasing \cite{wang2024grutopia, puig2023habitat, li2021igibson}.

\begin{figure*}[tb]
  \centering
  \includegraphics[width=\linewidth]{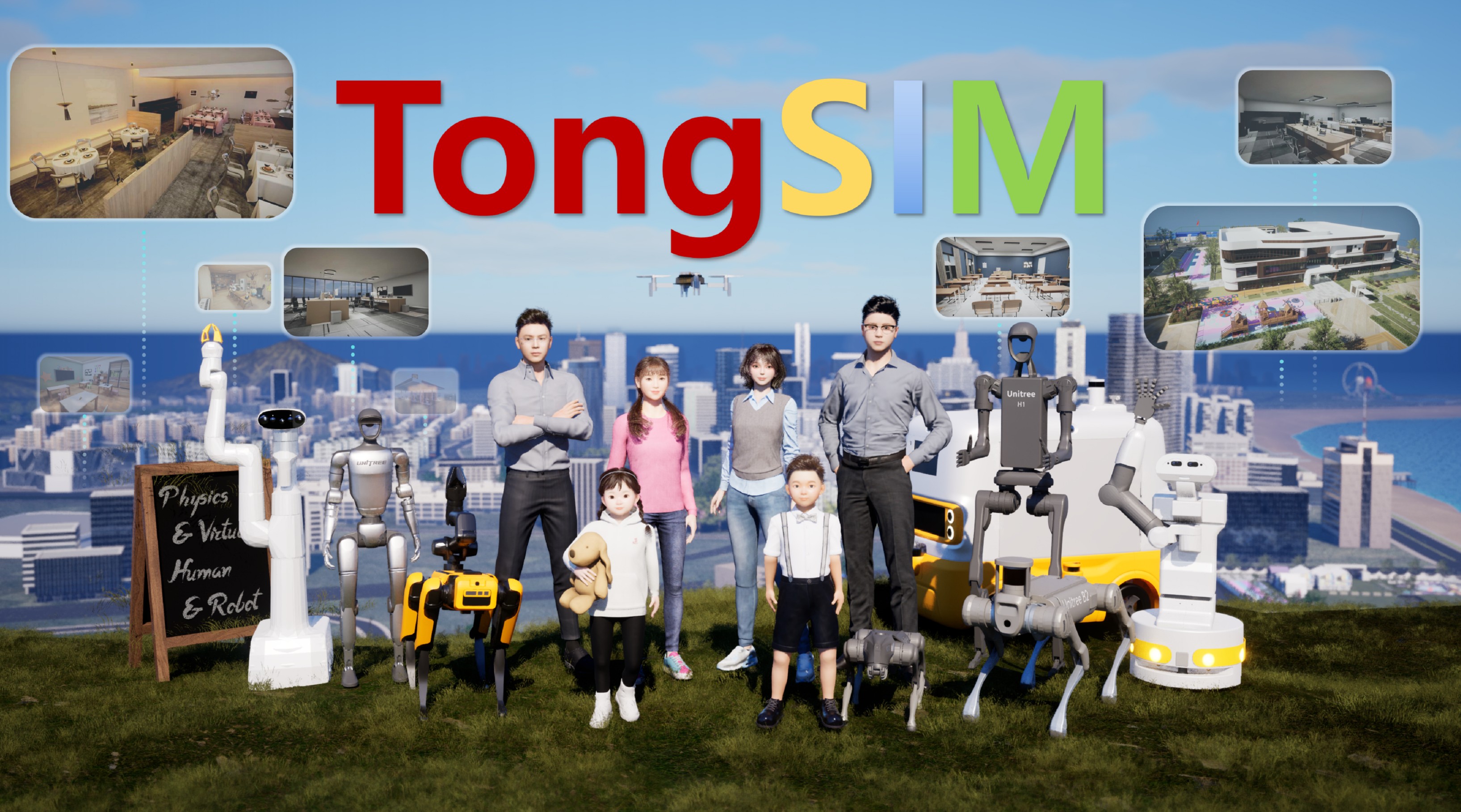}
  \caption{TongSIM, a simulation platform for general-purpose embodied AI agent training and evaluation. We provide diverse high-fidelity indoor and outdoor scenes that suit a large range of tasks, as well as multiple embodiments, including not only human-like figures but also robotic ones.}
  \label{fig:teaser}
\end{figure*}

Against this backdrop, a general-purpose simulation platform that can support both low-level tasks, such as embodied robot training, and high-level tasks, such as single-agent and even multi-agent social simulations, is essential. Such a platform can provide a highly consistent embodied general-purpose agent training environment, facilitating researchers from different fields to conduct model training, testing, and even develop new tasks on the same platform, which will help advance the development of general-purpose embodied intelligence.

In this work, we present \textbf{TongSIM}, a high-fidelity, universal embodied intelligence training and testing platform that supports complex indoor and outdoor scene simulation, as shown in Figure~\ref{fig:teaser}. In this platform, we have constructed a diverse range of indoor and outdoor scenes. 
Leveraging these diverse and rich scenes, we have proposed a series of benchmarks. These benchmarks cover a wide range of agent capabilities, including perception, cognition, decision-making, human-robot cooperation, spatial and social understanding, covering both low-level tasks (such as navigation) and high-abstraction tasks (such as multi-agent games), forming a comprehensive evaluation system that spans the entire spectrum \cite{peng2024tong}. Users can independently select different benchmarks to train or test specialized agents or integrate these benchmarks for general-purpose agents. Specifically, we propose 5 categories of benchmarking tasks, including single-agent tasks, multi-agent tasks, human-robot interaction tasks, primary family composite tasks, and advanced family composite tasks.

\section{Related Works}

Progress in Embodied AI is inextricably tied to developments in simulation technologies and evaluation benchmarks. High-fidelity simulation environments offer low-cost, safe, and reproducible training platforms for agents, while diverse benchmarks drive the evolution of these agents from foundational visual navigation to complex, long-horizon task execution.

\subsection{Simulators for Embodied AI}
The evolution of simulation platforms outlines a clear technical trajectory: transitioning from static, visually faithful environments designed for navigation, to physics-rich interactive worlds supporting fine-grained manipulation, and currently advancing towards generative, open-ended ecosystems powered by AI, aiming to reconcile the trade-off between training scalability and physical realism.

The Habitat platform \cite{savva2019habitat} epitomizes high-efficiency and large-scale simulation. Comprising the high-performance 3D simulator Habitat-Sim and the embodied reinforcement learning (RL) framework Habitat-Lab, its core advantage lies in extreme rendering throughput. Capable of running in parallel at thousands of frames per second, it significantly accelerates the training cycle of RL algorithms. Early iterations, primarily utilizing real-world 3D scan datasets (e.g., Matterport3D\cite{chang2017matterport3d}), focused on visual fidelity to support visual navigation tasks. With the release of Habitat 2.0 \cite{szot2021habitat}, the platform integrated the Bullet physics engine, extending capabilities to interactive navigation. Most recently, Habitat 3.0 \cite{puig2023habitat} introduced humanoid avatars, establishing ``Co-Habitat" scenarios for human-robot-environment coexistence, marking a shift towards social interaction capabilities. Sharing this focus on human-centric environments, VirtualHome \cite{puig2018virtualhome} utilizes Unity3D to abstract household activities into executable programs, specifically targeting action understanding and high-level task planning.

Unlike Habitat's emphasis on scene scale, SAPIEN \cite{xiang2020sapien} and iGibson 2.0 \cite{li2021igibson} prioritize interaction precision at the object level. SAPIEN focuses on manipulation tasks involving articulated objects, utilizing physics-driven active stereoscopic vision simulation to bridge the visual Sim-to-Real gap. iGibson 2.0 \cite{li2021igibson}, an open-source environment built on the PyBullet physics engine, emphasizes object-centric simulation. It supports various long-horizon household tasks, such as cleaning and cooking, and introduces an evaluation system based on logical states. 

To further narrow the Sim-to-Real domain gap, some simulators have adopted GPU-accelerated physics engines and ray-tracing technologies. NVIDIA's Isaac Sim \cite{makoviychuk2021isaac}, built on the Omniverse platform, leverages the PhysX 5 engine to provide high-fidelity physics simulation and integrates real-time ray tracing. It is widely applied in industrial robot manipulation and Sim-to-Real transfer research. As the successor to iGibson \cite{li2021igibson}, OmniGibson \cite{li2022behavior} has also migrated to the Omniverse architecture. Utilizing the PhysX backend, it achieves real-time simulation of complex materials such as fluids and cloth, accommodating large-scale benchmarks like BEHAVIOR-1K \cite{li2022behavior}.

\begin{table*}[tb]
\centering
\caption{Comparison of embodied AI simulators. We compare TongSIM (Ours) with state-of-the-art platforms across simulation engines, asset diversity, platform features, and supported tasks. TongSIM possesses the common features of current state-of-the-art simulators, and it stands out in city-level interaction, task-oriented fidelity, and the variety of the supported tasks.}
\label{tab:comparison}
\resizebox{\textwidth}{!}{
\begin{tabular}{l|l|c|c|c|c|c|c}
\toprule
\multicolumn{2}{c|}{\multirow{2}{*}{\textbf{Features}}} & \textbf{TongSIM} & \textbf{GRUtopia} & \textbf{OmniGibson} & \textbf{Habitat} & \textbf{VirtualHome} & \textbf{Virtual Community} \\
\multicolumn{2}{c|}{} & \textbf{(Ours)} & \cite{wang2024grutopia} & \cite{li2022behavior} & \cite{puig2023habitat} & \cite{puig2018virtualhome} & \cite{zhou2025virtual} \\ \midrule

\multirow{1}{*}{\textbf{Core}} & Engine Base & UE 5 & Isaac Sim & Isaac Sim & Bullet & Unity3D & Genesis \\  \midrule

\multirow{4}{*}{\begin{tabular}[c]{@{}l@{}}\textbf{Environment}\\ \textbf{\& Scenes}\end{tabular}} 
& Scene & 115 & 100 Annotated & 50  & 211 & 6 & 35 Urban Areas \\
& Indoor Scope & \cmarkg & \cmarkg & \cmarkg & \cmarkg & \cmarkg & \cmarkg \\
& Outdoor Scope & \cmarkg & \cmarkg & \cmarkg & \xmarkr & \xmarkr & \cmarkg \\
& City-level Interaction & \cmarkg & \cmarkg & \xmarkr & \xmarkr & \xmarkr & \cmarkg \\ 

\midrule

\multirow{4}{*}{\begin{tabular}[c]{@{}l@{}}\textbf{Platform}\\ \textbf{Features}\end{tabular}} 
& Parallel Training & \cmarkg & \cmarkg & \cmarkg & \cmarkg & \xmarkr & \xmarkr \\
& Task-oriented fidelity & \cmarkg & \xmarkr & \xmarkr & \xmarkr & \xmarkr & \xmarkr \\
& NPC Control & \cmarkg & \cmarkg  & \xmarkr & \cmarkg & \cmarkg & \cmarkg \\
& Sim-to-Real Support & \cmarkg & \cmarkg & \cmarkg & \cmarkg & \xmarkr & \cmarkg \\ \midrule

\multirow{3}{*}{\begin{tabular}[c]{@{}l@{}}\textbf{Supported}\\ \textbf{Tasks}\end{tabular}} 
& Single-Agent & \cmarkg & \cmarkg & \cmarkg & \cmarkg & \cmarkg & \cmarkg \\
& Multi-Agent & \cmarkg & \cmarkg & \xmarkr & \cmarkg & \cmarkg & \cmarkg \\
& Human-Robot Teaming & \cmarkg & \xmarkr & \xmarkr & \cmarkg & \xmarkr & \cmarkg \\
\bottomrule
\end{tabular}
}
\end{table*}

Furthermore, the field is evolving towards generative simulation. Genesis \cite{Genesis} acts as a nascent universal physics engine, unifying simulation frameworks for rigid bodies, fluids, and soft bodies, while exploring the use of generative AI to construct 4D dynamic worlds. Building upon this generative capability, Virtual Community \cite{zhou2025virtual} expands the simulation horizon from single households to city-level ecosystems. It leverages a generative pipeline to construct diverse urban districts, supporting complex physical and social planning tasks that require agents to navigate and interact within metropolitan contexts. Concurrently, ProcTHOR \cite{deitke2022️} employs procedural generation to automatically construct massive environments, offering an effective solution to the scarcity of scene data for training. We compared the features of TongSIM and several state-of-the-art simulators in Table~\ref{tab:comparison}.

Currently, simulation technology still faces a trade-off between runtime efficiency and physical fidelity: platforms pursuing extreme speed (e.g., Habitat \cite{savva2019habitat,szot2021habitat,puig2023habitat}) often simplify contact dynamics, whereas those prioritizing high-fidelity physics (e.g., Isaac Sim \cite{makoviychuk2021isaac}, OmniGibson \cite{li2022behavior}) incur significantly higher computational costs. Constructing a unified architecture that balances large-scale parallel training with high-precision physical interaction remains a core engineering challenge in the field.

\subsection{Benchmarks for Embodied AI}
Benchmarks in Embodied AI reflect the trajectory of task complexity: evolving from instruction following and navigation to object manipulation in household settings, and finally to general task planning in open worlds.

Early benchmarks primarily focused on an agent's spatial understanding of natural language instructions. Room-to-Room (R2R) \cite{anderson2018vision} is a seminal work in Vision-and-Language Navigation (VLN), requiring agents to plan paths in real-world scanned indoor environments based on linguistic instructions. Although R2R \cite{anderson2018vision} propelled the development of multimodal models, the simplicity of its action space and task state transitions limited its utility for evaluating complex embodied intelligence.

ALFRED \cite{shridhar2020alfred} elevated the challenge to long-horizon, compositional household manipulation tasks (e.g., ``rinse a mug and place it in the coffee maker"). ALFRED's core contribution lies in introducing irreversible state changes and long action sequences, rigorously testing an agent's long-term memory, task decomposition, and planning capabilities. This benchmark shifted the research focus from pure navigation to scenarios requiring fine-grained object manipulation, often grounded in interactive environments like AI2-THOR \cite{kolve2017ai2}.

Pursuing the goal of artificial general intelligence (AGI), subsequent benchmarks have sought greater scale and behavioral diversity. BEHAVIOR-1K \cite{li2022behavior} represents a significant step toward training generalist agents. Containing 1,000 everyday human activities, it emphasizes human-centric behaviors and realistic simulation, aiming to foster capabilities for diverse, open-ended tasks. Regarding environment scale, ProcTHOR \cite{deitke2022️} utilizes procedural generation to break the constraints of limited real-world scans, enabling the rapid creation of massive embodied AI environments.
After constructing simulation environments, the next step is to deploy tasks, which can be used for both evaluation and training. Recent research has increasingly emphasized dynamic task generation, rather than relying on traditionally manually collected tasks~\cite{deng2009imagenet, scene3d, Cui2025AbilityDA}. However, most task generation approaches are designed for highly structured domains, such as code-based tasks~\cite{zhao2025absolutezero}, GUI-based tasks~\cite{omnibench}, or game-based tasks~\cite{zheng2025mcu, tang2024mars}. Existing task generation approaches for embodied 3D simulation environments are largely limited to low-level variations, such as modifying scenes, objects, or spatial layouts~\cite{deitke2022️, wang2023robogen}.

Recent research explores ultra-long-horizon planning and larger-scale environments. CookBench \cite{cai2025cookbench} focuses on complex cooking scenarios, requiring intent recognition and fine-grained interaction over long sequences, posing severe challenges to the reasoning capabilities of LLMs and vision-language models (VLMs). GRUtopia \cite{wang2024grutopia} signifies a migration from indoor to city-scale environments, aiming to explore Scaling Laws in Embodied AI and providing the first simulation platform for general-purpose robots in complex, socialized urban scenarios.

These complex benchmarks collectively reveal a critical challenge: error accumulation in long-horizon planning. In tasks like CookBench, minor deviations in early actions can be amplified exponentially in subsequent steps, leading to task failure. Consequently, research into Embodied World Models is becoming increasingly vital, with the core challenge being the maintenance of temporal consistency and the mitigation of error accumulation. The evaluation paradigm is also shifting, moving beyond simple task success rates to quantifying an agent's capacity for physical consistency, state understanding, and counterfactual reasoning.

\section{TongSIM Platform}

\begin{figure*}[tb]
  \centering
  \includegraphics[width=\linewidth]{tongsim/figures/system_arch.pdf}
  \caption{Overview of the TongSIM system architecture. The platform contains a unreal engine-based simulator and a python controller which can communicate with the simulator. Within the simulator, four features are supported, including multimodal data sensors, high-fidelity simulation, large-scale NPC systems, and parallel training. Based on this platform, various tasks are developed to support embodied AI, such as indoor navigation, multi-agent cooperation, home composite tasks, etc. Also, the platform integrates the evaluation system to evaluate the performance of agents about tasks, abilities, and level of intelligence.}
  \label{fig:System Architecture}
\end{figure*}

\subsection{Platform Overview}
TongSIM is designed as a comprehensive and versatile simulation platform for general intelligent machines. Built upon Unreal Engine 5.6 (UE5.6), the platform extends the native capabilities of the engine through a suite of custom-developed wrappers and interfaces. These extensions facilitate robust communication, debugging, and control, effectively managing high-dimensional simulation data—including scene semantics, object states, and agent metadata—to streamline agent training, testing, and secondary development.

Specifically, the TongSIM platform comprises 115 distinct simulation scenarios, designed to support a diverse array of tasks. 
These environments are populated with thousands of high-fidelity objects.
To emulate realistic anthropomorphic behavioral patterns, we implemented 28 distinct interaction functions to satisfy the interaction requirements of both embodied agents and human users.

TongSIM provides rich semantic annotations to support learning tasks. We assign category labels to common objects, facilitating efficient filtering and supervised training. To support rule-based autonomous navigation for virtual humans and robots, we have baked navigation meshes into both indoor and outdoor environments. Furthermore, each environment is equipped with a Scene Manager, which provides users with ground-truth data, including scene metadata, segmentation maps, spawn points, and navigation information. All provided assets support custom secondary annotation, allowing researchers to tailor data to specific requirements.

The system architecture of TongSIM is illustrated in Figure \ref{fig:System Architecture}. To facilitate the integration of diverse agents and models, TongSIM provides a comprehensive application programming interfaces (APIs) of Python. This enables developers to exert granular control over the simulation environment and embodied agents via Python scripts. Key functionalities include: level management (loading/unloading), dynamic instantiation of objects and avatars, character manipulation, state retrieval, camera control, retrieval of objects near specific coordinates, and the execution of native UE console commands.

A typical operational workflow commences with the initialization of the TongSIM UE Server, followed by the deployment of a TongSIM UE Client to establish a connection. Utilizing the Python software development kit (SDK), users manipulate scenes, objects, NPCs, and agents, while the resulting visual states are rendered and streamed to a web interface for observation by both human operators and agents. Simultaneously, the TongSIM-Audio2Face module synthesizes and synchronizes facial expressions and vocalizations for NPCs and agents. Furthermore, the platform supports immersive interaction via virtual reality (VR) devices, requiring the execution of a dedicated VR Client to interface with the server.

\subsection{High Fidelity Scenes}
To support comprehensive and general-purpose agent training and testing, the TongSIM platform offers a vast and diverse collection of simulation environments featuring varying levels of fidelity. Specifically, this comprises 115 indoor environments and extends support to urban-scale outdoor scenarios.

\begin{figure*}[tb]
  \centering
  \includegraphics[width=\linewidth]{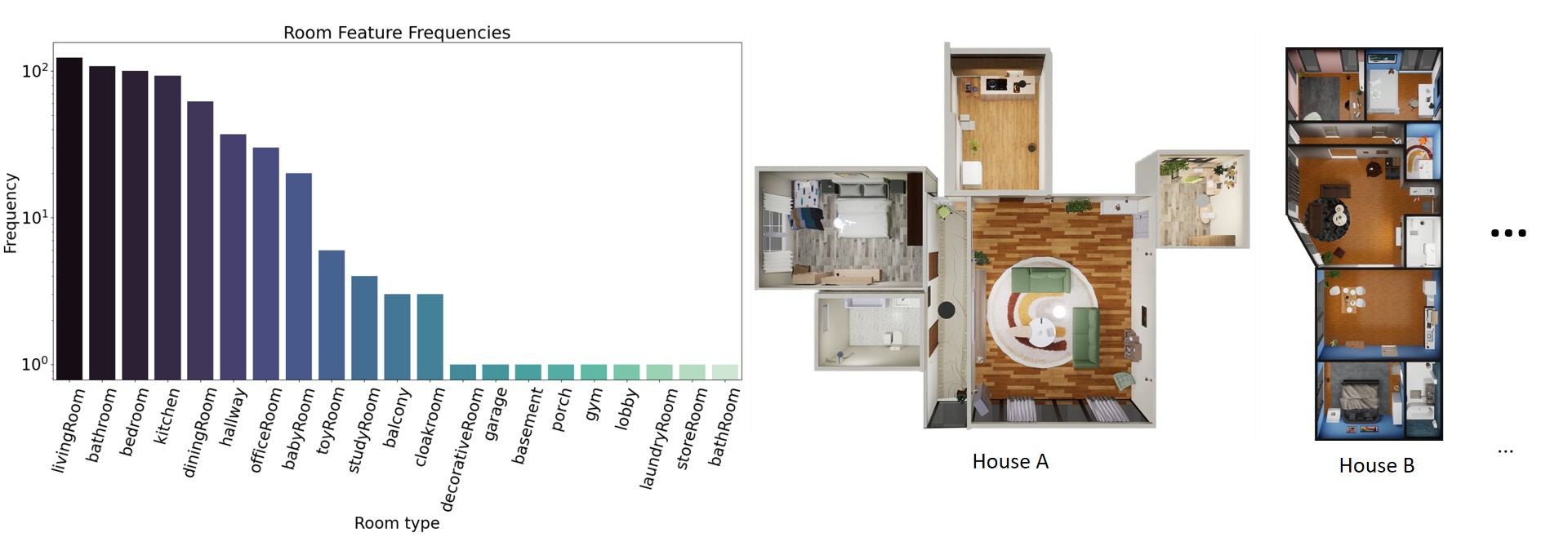}
  \caption{Statistics of indoor environments in TongSIM. The dataset spans diverse functional categories (e.g., residential, commercial) and architectural styles (e.g., modern, classical Chinese), designed to support complex, human-centric tasks.}
  \label{fig:indoor_stats}
\end{figure*}

Regarding indoor settings, a defining characteristic of TongSIM is its capacity to support complex hybrid daily living tasks. To ensure semantic realism, we engaged professional designers to manually curate spatial layouts and furniture arrangements, thereby guaranteeing that these environments strictly adhere to anthropomorphic behavioral logic. Building upon these expert-designed environments, we developed an automated expansion pipeline that scales the repository to a total of 115 distinct scenes. Functionally, these environments cover diverse categories such as residential units, cafes, and retail stores. Stylistically, they span a wide architectural spectrum, ranging from modern apartments and villas to medieval castles, traditional Japanese gardens, and classical Chinese architecture. Detailed statistics regarding the indoor scenes are presented in Figure \ref{fig:indoor_stats}.

\begin{figure*}[tb]
  \centering
  \includegraphics[width=0.95\linewidth]{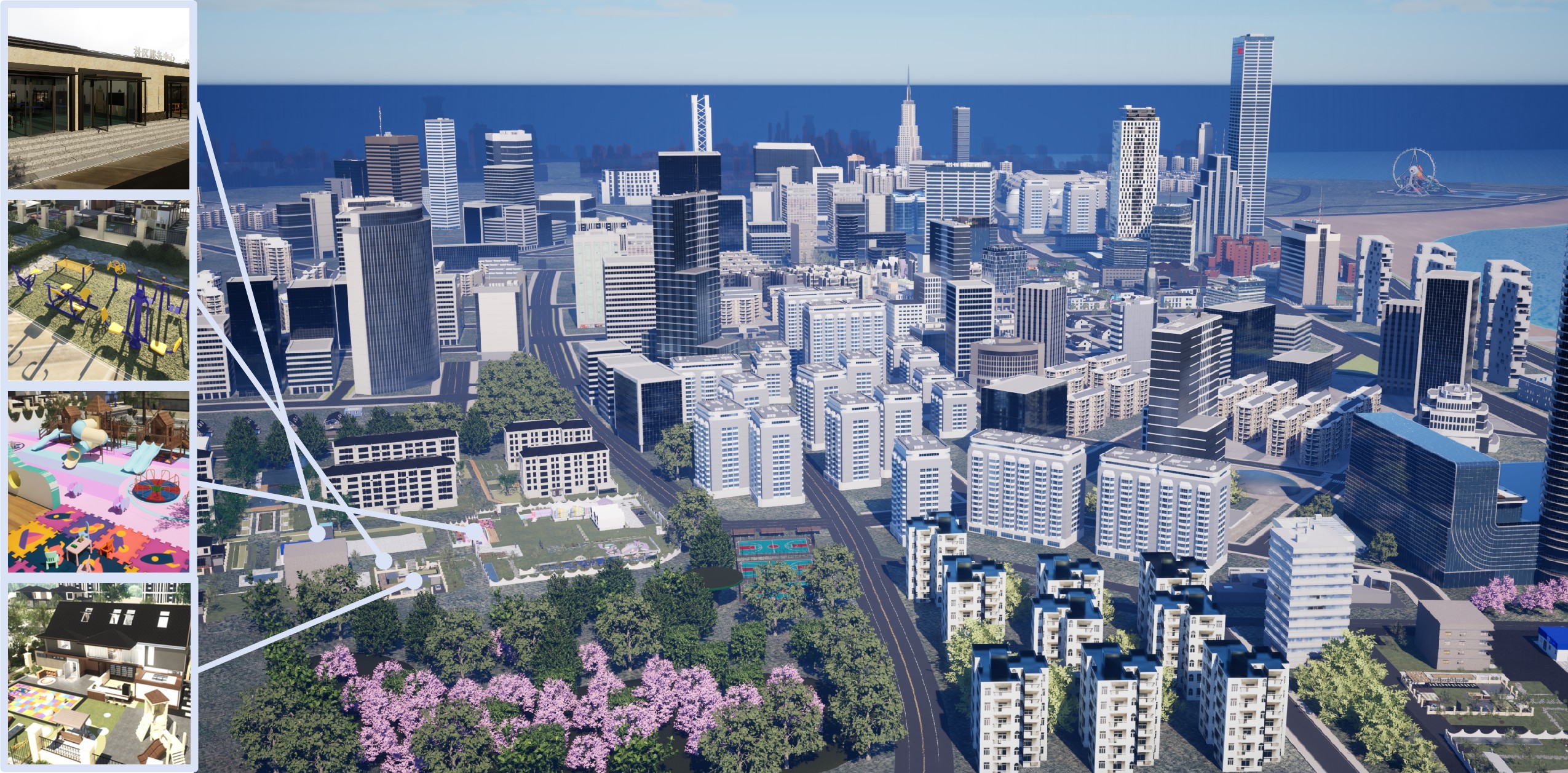}
  \caption{Visualization of the TongSIM Outdoor World. The platform simulates a holistic, spatially contiguous urban environment rather than isolated fragments.}
  \label{fig:outdoor_world}
\end{figure*}

Regarding outdoor scenarios, rather than assembling isolated environmental fragments, we have constructed a holistic virtual metropolis, as illustrated in Figure \ref{fig:outdoor_world}. This unified world encompasses diverse functional zones, including educational institutions, residential complexes, commercial districts, and medical facilities, along with a comprehensive road network and a dynamic traffic simulation system. Crucially, these functional zones are spatially contiguous. This design enables embodied agents to navigate seamlessly across distinct regions, thereby preserving contextual continuity during long-horizon training and evaluation tasks.

\subsection{Customize Tasks and Scenes}
To enhance the scalability and diversity of the TongSIM environment, we developed an automated procedural generation pipeline capable of expanding the scene repository based on existing assets. This method follows a coarse-to-fine strategy.
In the coarse-grained phase, we decompose existing indoor scenes into independent functional units (e.g., bedroom, study, kitchen). These units are then stochastically recombined to synthesize novel layout configurations, where the door frames serve as alignment anchors.
In the fine-grained phase, we introduce micro-level variations to the interior. Specifically, we apply random perturbations to the poses of movable objects and randomly replace a subset of assets with the same-category counterparts retrieved from our object library. This process significantly enriches the diversity of interior decor, spatial layouts, and object appearances.
Finally, to guarantee the high fidelity of the TongSIM platform, we employ a human-in-the-loop validation step where professional designers filter and fine-tune the generated scenes, pruning any instances that exhibit semantic implausibility.

Complementing the automated generation pipeline, TongSIM features a robust external content import framework. This framework is designed for high compatibility with mainstream digital content creation (DCC), computer-aided design (CAD), and 3D scanning workflows, allowing existing assets to be seamlessly integrated into the platform’s simulation tasks and evaluation pipelines. Through a minimal preprocessing procedure, including comprising unit standardization, coordinate system alignment, and lighting configuration, users can efficiently import custom environments or third-party assets. Currently, TongSIM supports formats including glTF 2.0 (.gltf/.glb), FBX Scene, and Datasmith (.udatasmith), comprehensively covering industry-standard 3D data representations.

\subsection{Agent}

TongSIM provides diverse agents that can serve both as embodiments for AI models during training and evaluation, and as NPCs to facilitate task execution and enhance environmental fidelity. To govern the behaviors of these NPCs, TongSIM implements a hybrid automatic control mechanism driven by rule-based logic and LLMs.

TongSIM equips these agents with versatile functional capabilities and diverse visual appearances, allowing external AI models to drive embodiments via a Python API. The supported action space spans multiple levels of complexity, including kinematic primitives (e.g., nodding, waving, turning), target-driven behaviors directed at specific coordinates or objects (e.g., gazing, point-to-point navigation), and fundamental object interactions (e.g., pick-and-place, toggling doors, sitting/standing). Furthermore, the platform supports complex composite activities that involve multi-step sequencing and high-level semantics, such as consuming items, pouring liquids, mopping, wiping surfaces, reading, cutting food, and simulating daily routines like sleeping or washing.

\subsection{Platform Features}
\subsubsection{Physical Simulation}
The platform leverages the built-in Chaos physics engine of Unreal Engine 5 to achieve rigid body dynamics, fluid simulation, destruction, cloth, etc. This feature makes TongSIM suitable for constructing relative complicated 3D scenes, which supports the testing for embodied AI like robots. Meanwhile, we also try to integrate another physics simulation library (NVIDIA Flex) into TongSIM, which uses a unified particle to represent all object types, allowing different simulation materials to seamlessly interact.

\subsubsection{Interactive Objects}
\begin{figure*}[tb]
  \centering
  \includegraphics[width=\linewidth]{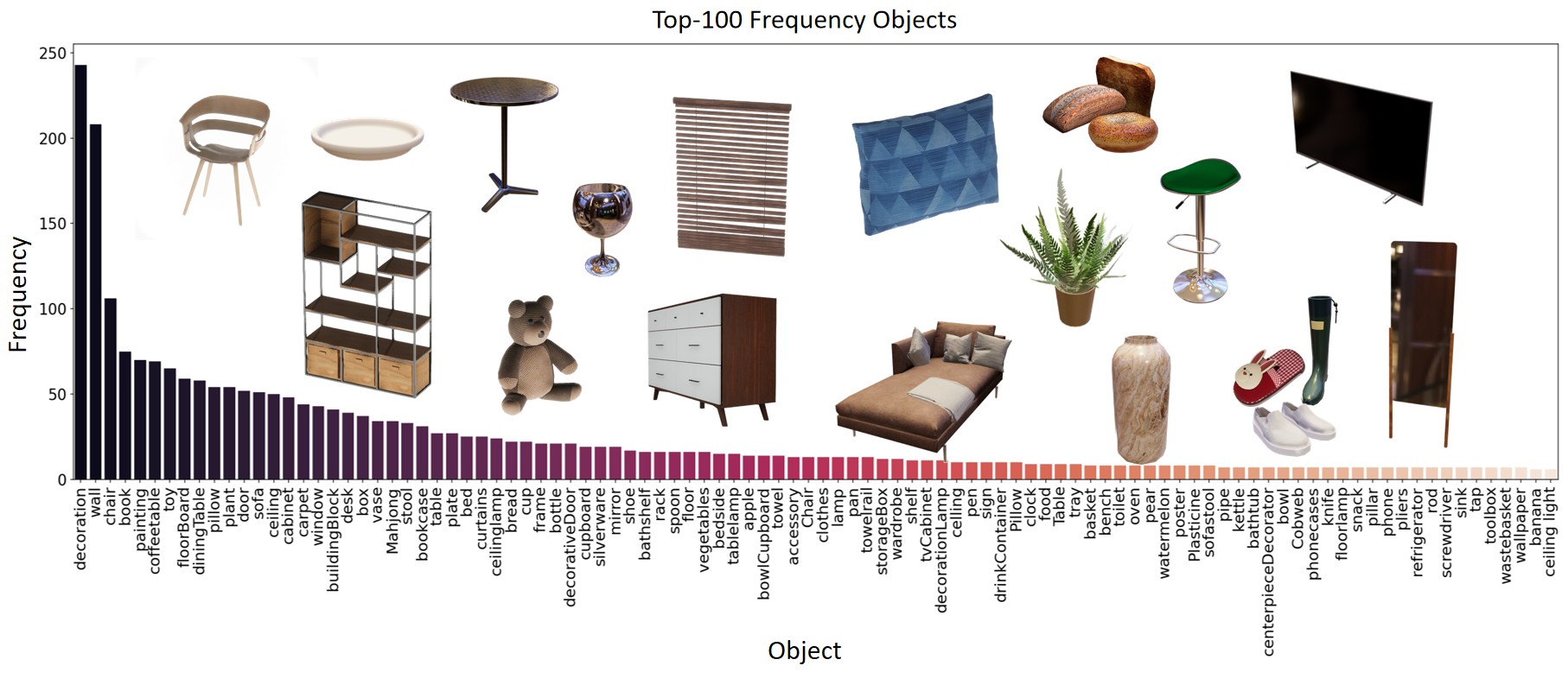}
  \caption{Statistics of the objects in TongSIM. 
  The dataset spans diverse functional categories (e.g., residential, commercial) and architectural styles (e.g., modern, classical Chinese), designed to support complex, human-centric tasks.}
  \label{fig:object_stats}
\end{figure*}

The platform constructs a protocol-based interaction system, achieving granular control over simulation entities through core protocols such as interactable ability. The platform supports 28 distinct interaction primitives and provides thousands of interactable objects. The statistics of some of the objects in TongSIM are shown in Figure~\ref{fig:object_stats}. This system not only covers basic operations, such as entity lifecycle management (generation, destruction) and geometric transformations (rotation, scaling), but also implements deep simulation of electromechanical logic and spatial semantic.

\textbf{Electromechanical Logic.} The system employs a dual-layer control mechanism that decouples the ``powered state'' from the ``activation state'' to accurately simulate the operational principles of real-world devices. The final operational state of a device is jointly determined by its physical power supply and switch control status; consequently, if an entity lacks a power connection, it remains non-functional even if logically activated. This design faithfully reproduces the physical dependency between electrical connectivity and device manipulation.

\textbf{Spatial Interaction Anchors.} To address the challenges of fine-grained manipulation in complex environments, the platform defines semantically annotated spatial interaction anchors. These anchors guide end-effectors (e.g., hands, robotic arms) to precisely target the coordinates of functional zones, such as handles or buttons, thereby establishing robust connection constraints. Furthermore, to facilitate composite interactions involving multi-object synergy, where task completion relies on specific spatial coordination between entities, the system introduces the concept of Placement Points. These points serve as predefined spatial docking slots that automatically calibrate the relative alignment between objects, such as aligning a cup beneath a water dispenser.


\begin{figure*}[tb]
  \centering
  \includegraphics[width=0.6\linewidth]{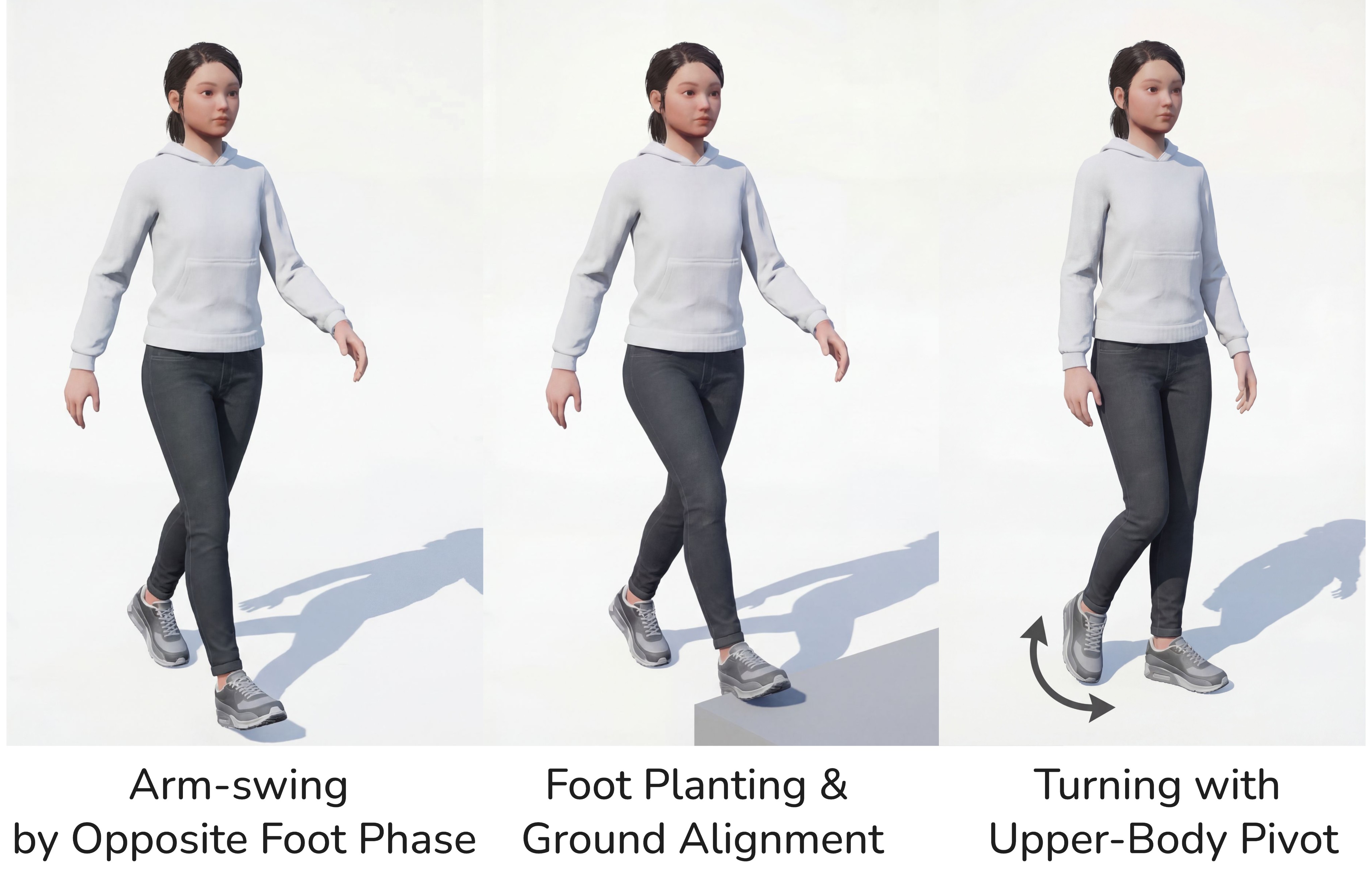}
    \caption{In TongSIM, programmatic animation functions, as a unified fundamental motion layer, are integrated with the platform's task, navigation, physics, and perception pipelines. Programmatic animation supports parameterization and controllable randomness, maintaining stability across different scenes. Users can rapidly deploy it in any scene and benchmark task without needing to modify the skeleton assets.}
    \label{fig:automotion}
\end{figure*}

\subsubsection{Experimental features}
\textbf{Automatic Procedural Actions.} This module provides animation-asset-free, procedural locomotion and arm articulation capabilities for humanoid agents within the TongSIM platform. An example is shown in Figure~\ref{fig:automotion}. Leveraging Control Rig and Inverse Kinematics (IK), the system synthesizes naturalistic gaits in real-time, adapting seamlessly to diverse skeletal structures and proportions. It is designed for real-time, large-scale interactive scenarios. 
In TongSIM, procedural animation serves as a unified foundational locomotion layer, deeply integrated with the platform’s task, navigation, physics, and perception pipelines. By supporting parameterization and controllable stochasticity, the module ensures stability across diverse scenes, allowing users to rapidly deploy agents across arbitrary environments and benchmark tasks without modifying skeletal assets.

\begin{figure*}[tb]
  \centering
  \includegraphics[width=\linewidth]{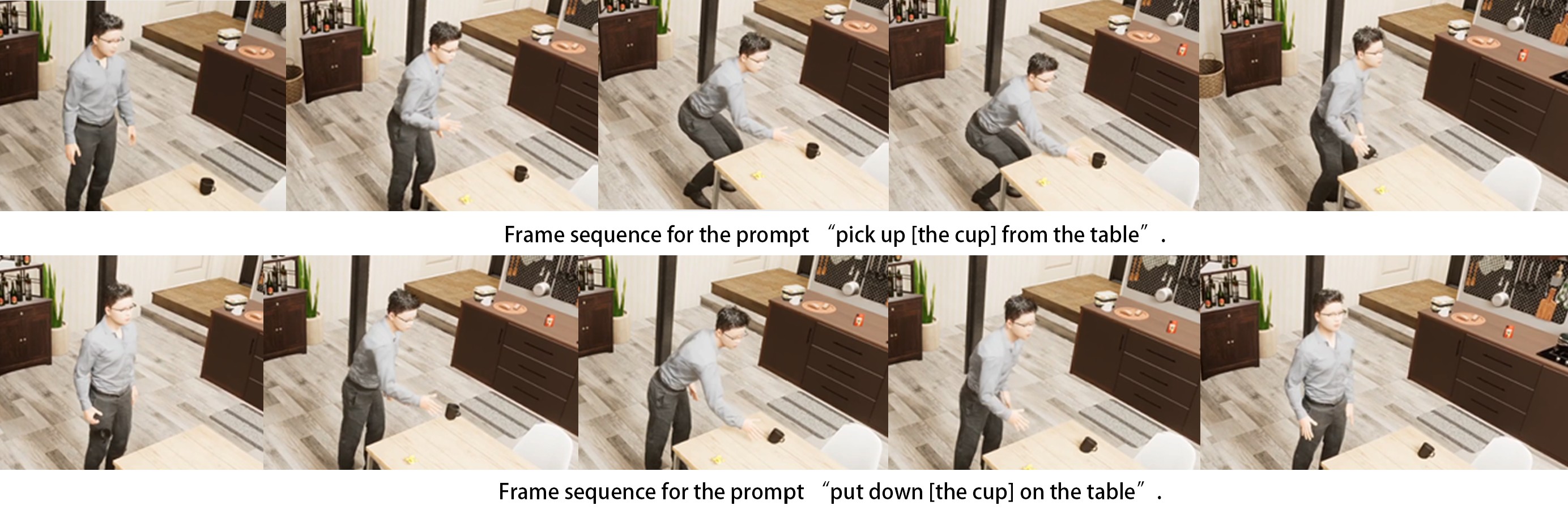}
  \caption{A typical procedure of text-driven motion generation.}
  \label{fig:text_driven_motion}
\end{figure*}

\textbf{Text-Driven Motion Generation.}
This module implements a diffusion-based motion generation model that synthesizes playable character animation sequences in real-time, utilizing textual intent and environmental voxel maps as inputs. A typical procedure is shown in Figure~\ref{fig:text_driven_motion}. The system features the following core capabilities.
(1) Text-driven control (natural language + target context): The system encapsulates natural language expressions alongside target positions and interaction objects within instructions. These are unified and parsed into executable intents and constraints for the model.
(2) Environmental perception (voxelized geometric semantics): The system performs real-time sampling of voxel grids around the character's current position and the first two trajectory waypoints. This constructs spatial semantics representing navigable areas and obstacles to facilitate model planning and collision avoidance.
(3) Segmented generation and streaming playback: Leveraging a bi-directional gRPC streaming pipeline, the system implements a ``generate-while-playing'' mechanism. This ensures seamless concatenation and automatic continuation of subsequent sequences, achieving low-latency real-time generation. 
(4) Voxel generation engine: This component provides comprehensive environmental voxel encoding. It constructs a voxel volume around the first two path waypoints, performs parallel sampling of geometric occupancy, and generates a bitmap byte stream transmitted with the request.
(5) Controllable root motion fusion: The server dictates the rhythm of vertical motion (e.g., undulation and footfall timing), while horizontal displacement and steering are dynamically planned by local navigation and control modules.

\begin{figure*}[tb]
  \centering
  \includegraphics[width=\linewidth]{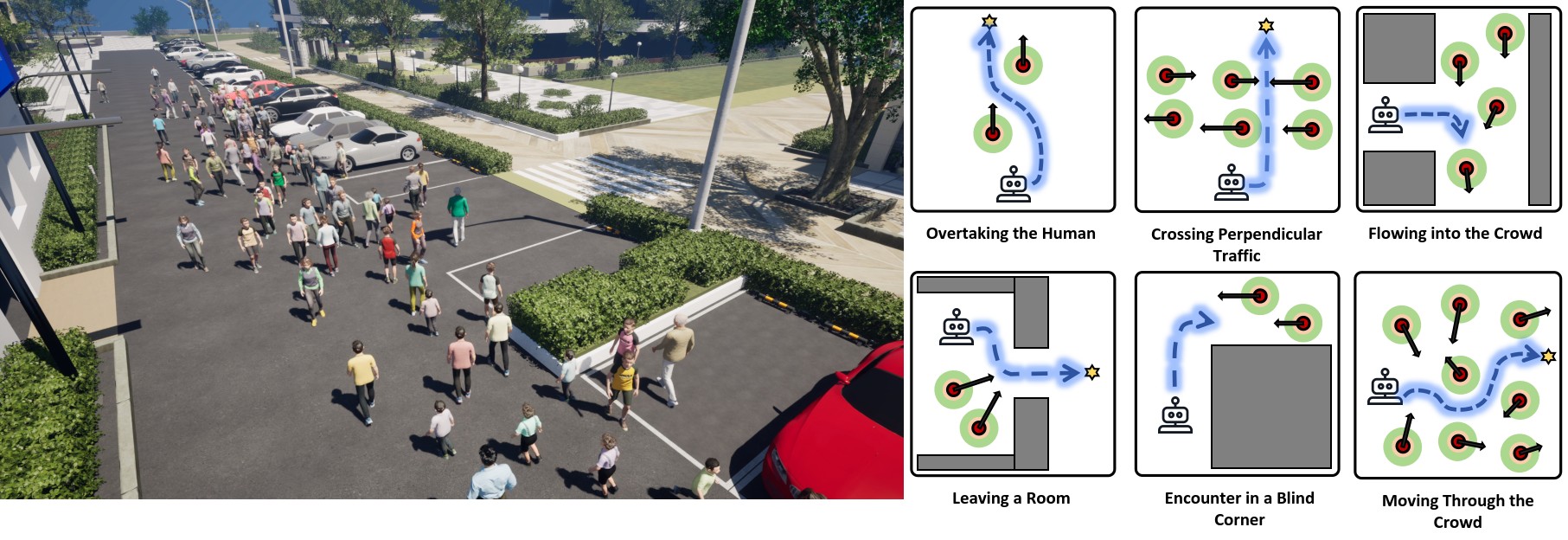}
  \caption{Simulation of diverse crowd and pedestrian movement patterns.}
  \label{fig:npc_crowd_motion}
\end{figure*}

\textbf{Large-Scale Crowd Simulation.}
To support human-robot hybrid tasks (e.g., social navigation, collaborative guidance, and interactive behavior learning), as shown in Figure~\ref{fig:npc_crowd_motion}, we constructed a hierarchical crowd simulation module designed to simultaneously achieve individual-level physical fidelity and group-level behavioral diversity.
(1) Low-level motion control: Social force model (SFM) and A*-based feasible region sampling. The foundational layer employs the social force model, introducing force-based interpersonal interactions and environmental constraints to realize natural obstacle avoidance and aggregation behaviors in local space.
(2) High-level decision making and planning: At the high-level decision layer, the system incorporates VLMs to simulate the semantic behaviors of human agents and support interaction and collaboration with robotic agents. This mechanism endows crowd simulation with semantic controllability and task relevance, achieving semantic alignment and intelligent synergy with the robot's task planning modules.

\textbf{From Simulation to Reality.}
To facilitate realistic robotic tasks and bridge the Simulation-to-Reality (Sim-to-Real) gap, we have explored two experimental integration strategies.
The first strategy involves replacing the native UE5 physics engine with the open-source MuJoCo engine \cite{todorov2012mujoco}. In this configuration, the physical simulation of all scene entities is delegated to MuJoCo, and the resulting state updates are synchronized with UE5’s rendering pipeline for visual output. This approach integrates MuJoCo’s superior physical fidelity and robust robotic training capabilities into TongSIM, effectively compensating for the limitations of UE5’s native dynamics simulation. The second strategy introduces native support for Isaac Lab, migrating both the physics and rendering engines to Isaac Sim. Crucially, this integration preserves TongSIM’s established task architecture and agent interfaces. This design facilitates seamless alignment with the robotics community, allowing for the reuse of extensive toolchains within the robotics research ecosystem. Both approaches are currently in the experimental phase, aiming to extend TongSIM’s compatibility to a broader spectrum of agent types and achieve a unified simulation framework for diverse embodiments.

\begin{figure*}[tb]
    \centering
    \begin{subfigure}[b]{0.46\linewidth}
        \centering
        \includegraphics[width=\textwidth]{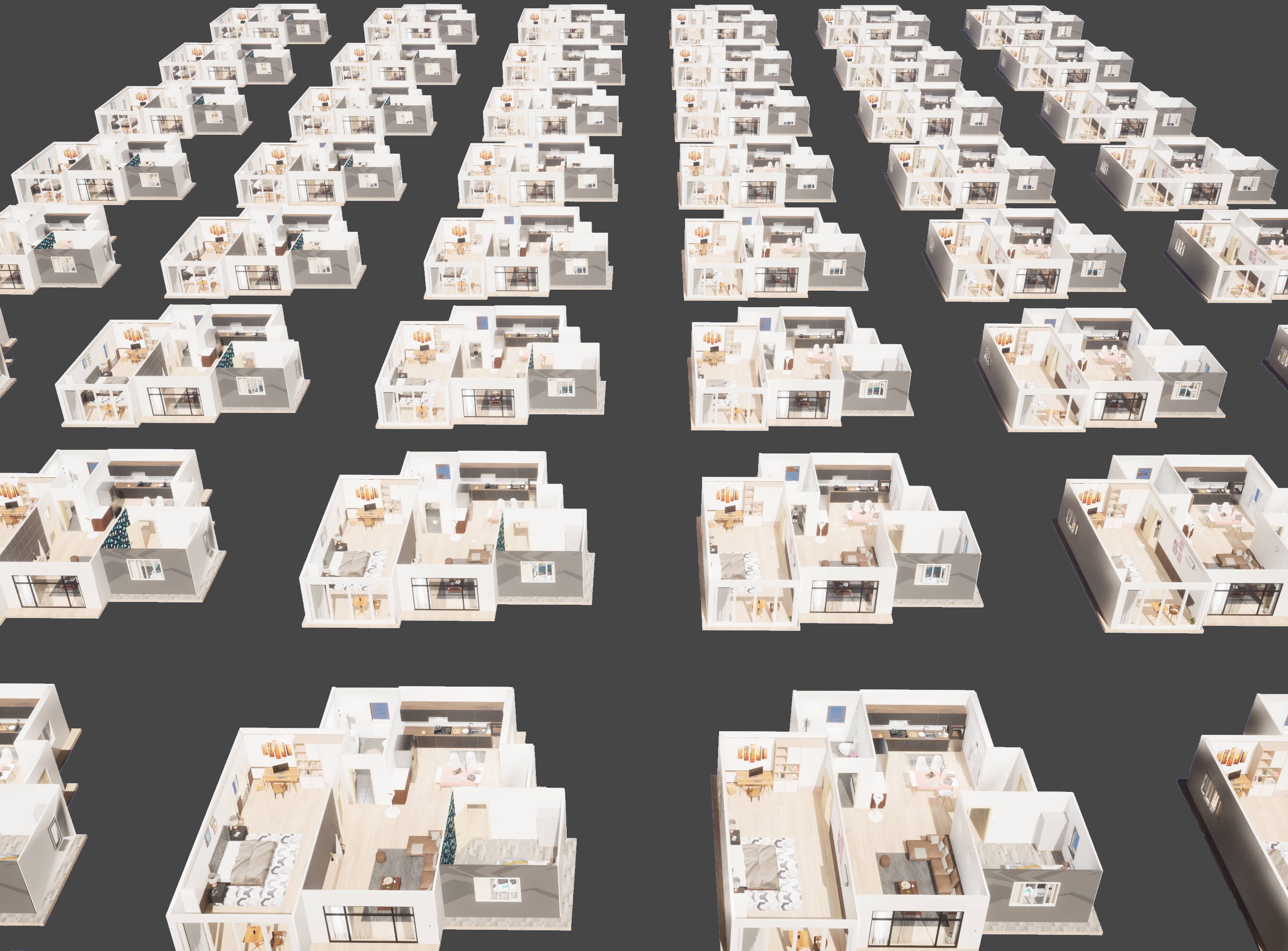}
        \caption{}
        \label{fig:parallel_training_a}
    \end{subfigure}
    \hfill
    \begin{subfigure}[b]{0.46\linewidth}
        \centering
        \includegraphics[width=\textwidth]{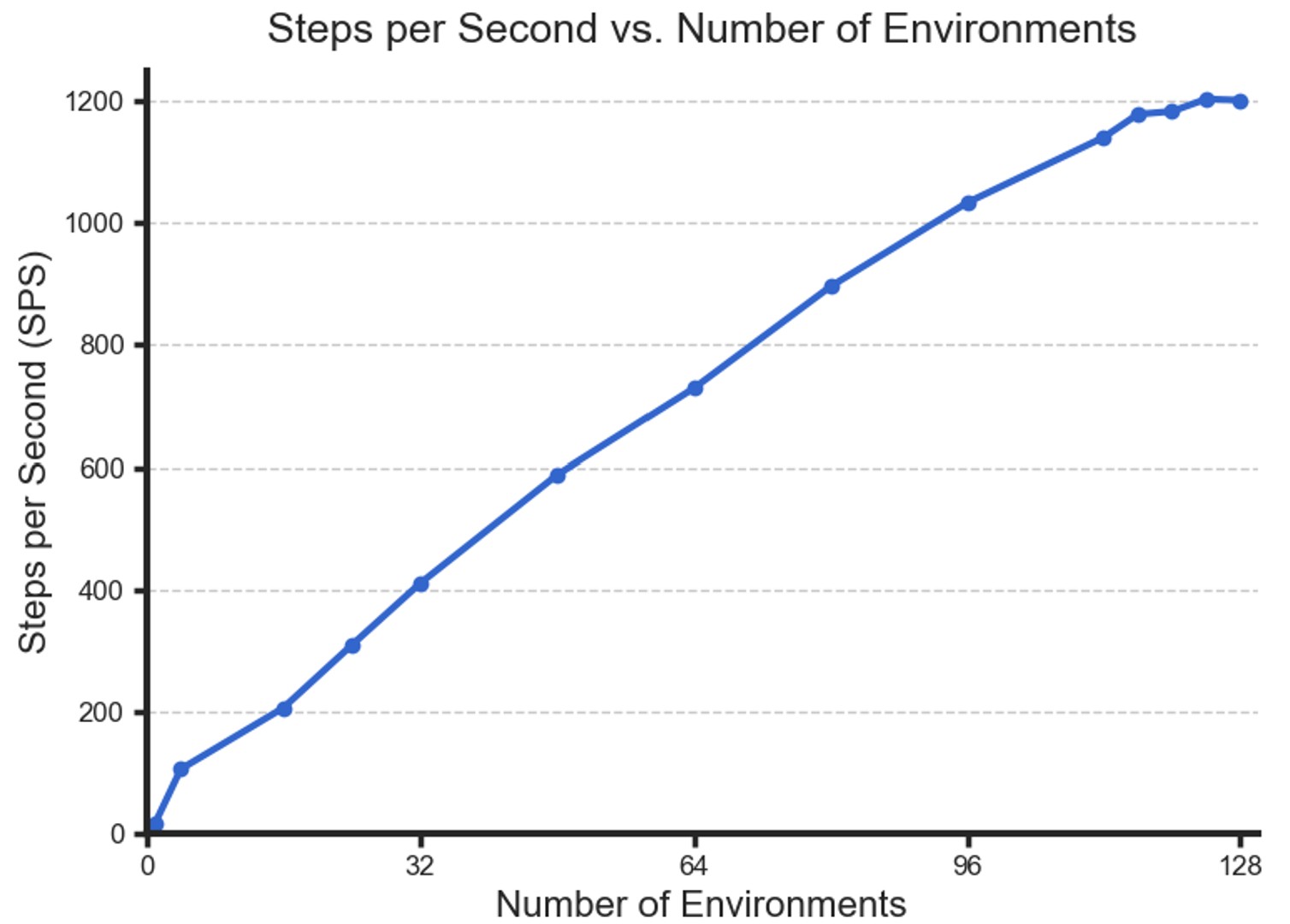}
        \caption{}
        \label{fig:parallel_training_b}
    \end{subfigure}
    \caption{Parallel training. (a) Visualization of 48 Parallel Environments.  (b) The steps per Second varies as the number of parallel environments.}
    \label{fig:parallel_training}
\end{figure*}

\subsubsection{Parallel Training}
To accommodate the demands for efficient agent training and optimize data sampling efficiency in RL, TongSIM introduces a multi-environment parallel execution mechanism. By simultaneously loading multiple mutually independent sub-levels within a single UE instance, this mechanism enables agents to concurrently acquire interaction data from diverse environments at each time step, thereby significantly amplifying sample generation throughput.

To evaluate the performance of this mechanism, we utilized the ``spatial exploration and navigation" task (see Subsect.~\ref{bench_1}) as a benchmark. Experiments were conducted on a personal workstation equipped with an Intel Core i9-13900KF CPU (24 cores/32 threads, 3.00 GHz) and an NVIDIA GeForce RTX 4090 GPU. Experimental results demonstrate that, compared to single-environment sequential execution, the parallel setting achieves a substantial increase regarding the interaction throughput measured in steps per second. Specifically, the sampling rate exhibits near-linear scaling as the number of parallel environments increases, as shown in Figure~\ref{fig:parallel_training} (b). However, at higher degrees of parallelism, performance gains gradually saturate due to overhead associated with inter-process communication and system scheduling.

TongSIM’s parallel execution mechanism effectively enhances sample acquisition efficiency and minimizes training time costs, establishing a robust foundation for large-scale reinforcement learning in complex simulation environments.

\section{Benchmarks}

\subsection{Single Agent Task: Spatial Exploration and Navigation} \label{bench_1}

\subsubsection{Benchmark Overview} 
Autonomous exploration and navigation represent critical pillars for intelligent agents operating in partially observable physical environments. Effective exploration demands that an agent maximize coverage of the state space, converting unknown areas into semantic maps to locate targets of interest. Subsequently, navigation entails executing optimal paths to reach these identified locations while negotiating obstacles. To rigorously evaluate these competencies, we introduce the spatial exploration and navigation test, a benchmark specifically designed to assess the exploration and navigation performance of intelligent agents. 

\begin{figure}[t]
  \centering
  \includegraphics[width=0.7\linewidth]{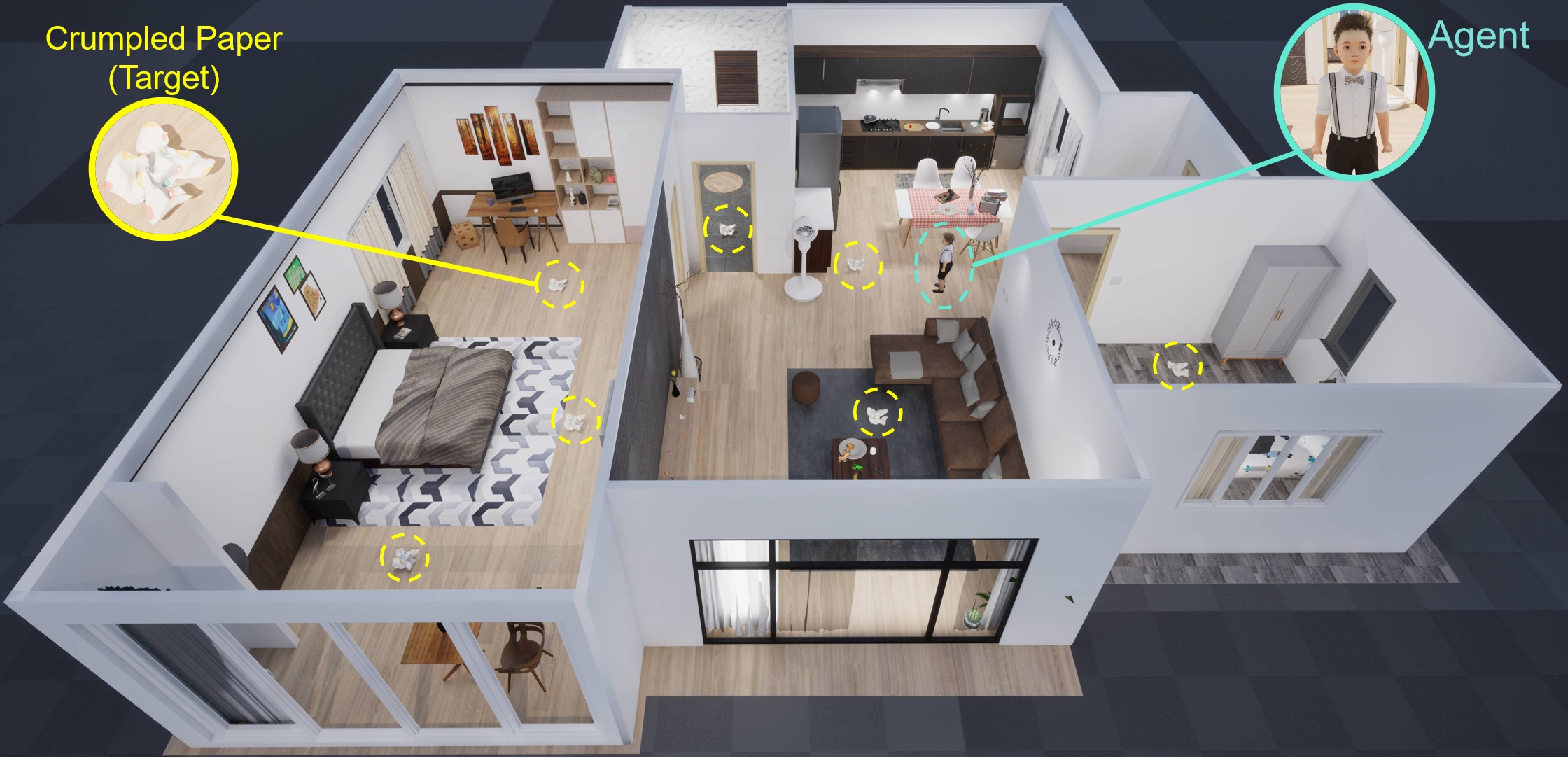} 
  \caption{An exemplary scenario of the paper ball cleaning task. The environment consists of multiple rooms, each furnished with various items, with paper balls scattered randomly across the floor. The agent under test is spawned at a random location within one of the rooms. It is required to explore these rooms to clean up all the paper balls.}
  \label{fig:paper_scenarios}
\end{figure}

The benchmark centers on a challenging cleanup task,as illustrated in Figure \ref{fig:paper_scenarios}, requiring the agent to navigate complex, multi-room indoor environments cluttered with obstacles to collect scattered paper balls. To foster diversity, the quantity and spatial distribution of the targets, as well as the agent's initialization pose, are fully randomized. Crucially, to guarantee task feasibility, we pre-calculate the traversable free space based on the agent's collision geometry and environmental obstacles, ensuring that both the agent and all targets are strictly spawned within reachable regions. The environment provides a rich observation space comprising egocentric RGB images, depth maps, and voxel grids. Correspondingly, the agent utilizes an action space designed for navigation, enabling movement to specific coordinates and rotation toward target orientations.

\subsubsection{Baseline}
In this benchmark, we project the volumetric voxel data of the task space onto a 2D plane to generate an occupancy grid. Specifically, we extract an egocentric $19 \times 19$ local grid, corresponding to a physical area of $208\ \text{cm} \times 208\ \text{cm}$, to serve as the agent's observation. To establish a robust baseline for comparison, we train a policy using proximal policy optimization (PPO) algorithm \cite{schulman2017proximal}. Furthermore, we incorporate human trials into the evaluation framework to quantify the performance gap between the autonomous agent and human operators.

\subsubsection{Metrics}

We employ two primary metrics to evaluate agent performance: \textbf{success rate (SR)} and \textbf{efficiency}.
The success rate measures the ratio of episodes where the agent successfully explores the entire task space and eliminates all scattered debris within a predefined maximum number of steps ($T_{\text{max}}$). 
The efficiency quantifies the agent's temporal performance based on step consumption. Since fewer steps indicate superior performance, we formulate efficiency as a normalized score relative to the maximum allowable steps.
The success rate is calculated as follows:
\begin{equation}
    \text{SR} = \frac{N_{\text{success}}}{N_{\text{total}}} \times 100\%
\end{equation}
where $N_{\text{total}}$ represents the total number of evaluation episodes, and $N_{\text{success}}$ denotes the number of successful episodes.
The efficiency is defined as:
\begin{equation}
    \text{Efficiency} = \frac{1}{N_{\text{success}}} \sum_{i=1}^{N_{\text{success}}} \left( \frac{T_{\text{max}} - S_i}{T_{\text{max}}} \right)
\end{equation}
where $N_{\text{success}}$ is the count of successful episodes, $T_{\text{max}}$ represents the maximum number of steps allowed per episode, and $S_i$ denotes the number of steps taken by the agent to complete the $i$-th successful episode.
\subsubsection{Experiments and Results}

\begin{table}[h]
    \centering
    \caption{Performance comparison between Human and RL Agent (PPO) on the benchmark.}
    \label{tab:single_agent_performance}
    \begin{tabular}{lcc} 
        \toprule
        \textbf{Agent} & \textbf{Success Rate} & \textbf{Efficiency} \\
        \midrule
        PPO   & 0.6 & 0.34 \\
        Human & 1.0 & 0.54 \\
        \bottomrule
    \end{tabular}
\end{table}

Using the proposed benchmark, we evaluated both the RL baseline model (PPO) and human operators. The comparative results are presented in Table \ref{tab:single_agent_performance}. Quantitative analysis reveals that the RL agent achieves a success rate of only 60\%. While the agent demonstrates the capability to complete the cleanup task to a certain extent, its performance significantly lags behind the human baseline in both success rate and efficiency.

To investigate the underlying causes of the agent's suboptimal performance, we conducted a qualitative analysis of the failure cases. We identified two primary limitations as follows. (1) Obstacle avoidance in cluttered environments: The RL agent frequently collides with obstacles or becomes trapped within narrow, obstacle-dense regions, indicating a deficiency in fine-grained collision avoidance capabilities. (2) Long-horizon navigation: The agent exhibits difficulty in traversing between distinct rooms, highlighting limitations in its spatial planning and long-horizon navigational reasoning.

\subsection{Multi Agent Task: Multi-Agent Cooperative Search}

\subsubsection{Benchmark Overview}
\begin{figure*}[tb]
    \centering
    \begin{subfigure}[tb]{0.72\linewidth}
        \centering
        \includegraphics[width=\textwidth]{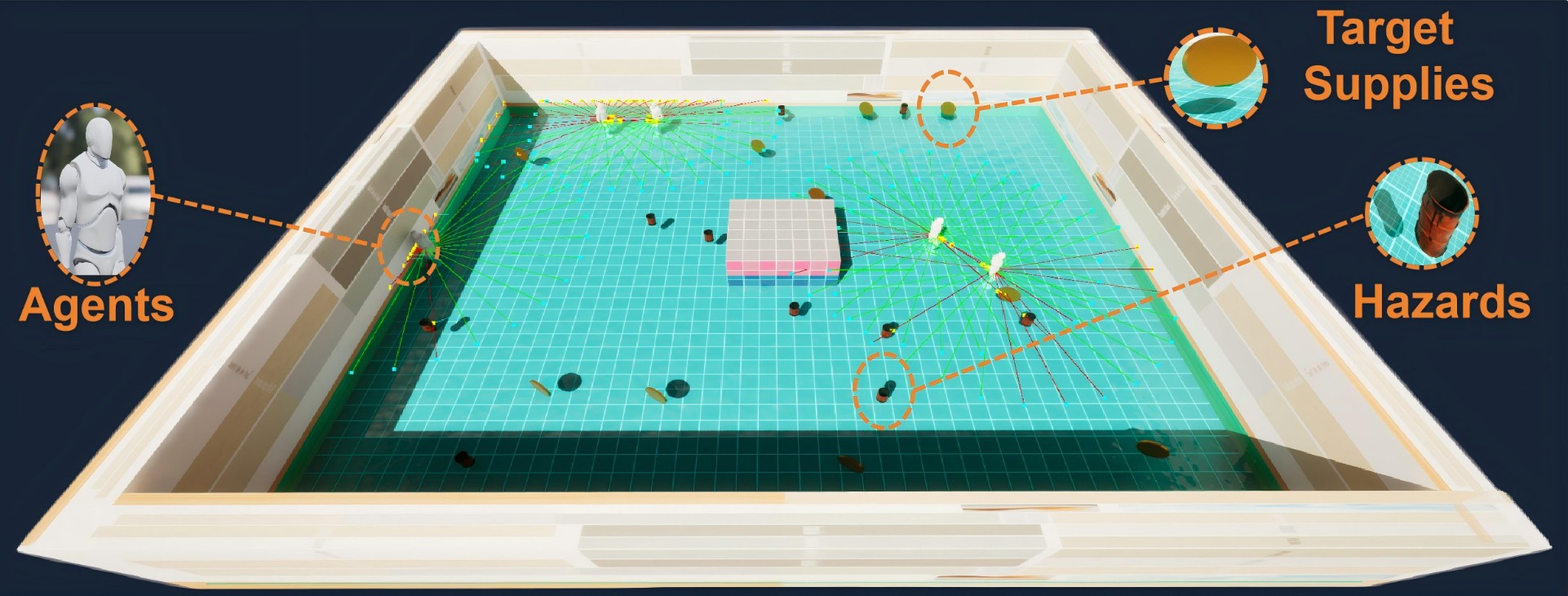}
        \caption{}
        \label{fig:macsr_a}
    \end{subfigure}
    \hfill
    \begin{subfigure}[tb]{0.26\linewidth}
        \centering
        
        \includegraphics[width=\textwidth,height=3.877cm]{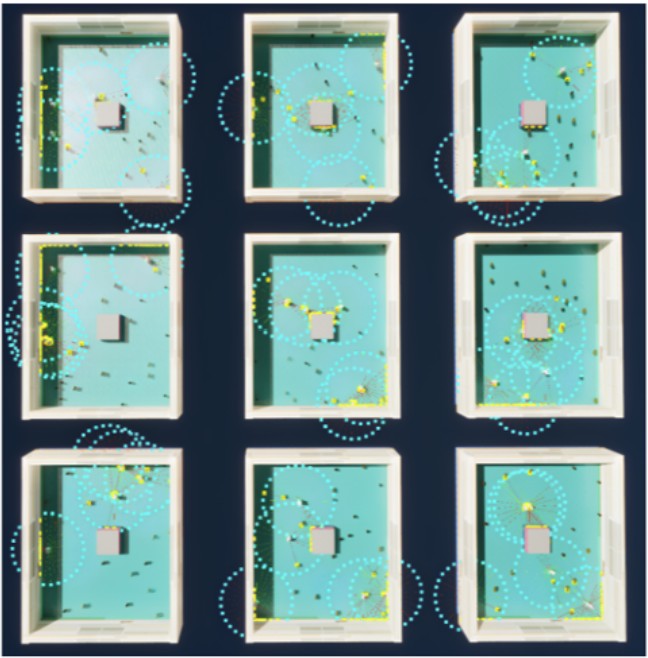}
        \caption{}
        \label{fig:macsr_b}
    \end{subfigure}
    \caption{Visualization of the multi-agent cooperative search (MACS) task scenarios. (a) Detailed scenario view. Key entities are explicitly labeled: agents, target supplies requiring collaborative collection, and dynamic hazards that must be proactively evaded. The green lines emanating from the agents provide a visualization of the radial ray-casting sensor array, illustrating the localized perception mechanism under conditions of partial observability. (b) Parallel training array: The top view of $3\times3$ array showing concurrent execution environments utilized for scalable multi-agent training. The blue dotted circumferences delineate the maximum sensory detection range for individual agents within the complex environment.}
    \label{fig:macsr_scenarios}
\end{figure*}

To facilitate research on multi-agent collaboration in complex 3D environments, we introduce the multi-agent cooperative search (MACS) task. Built upon the TongSIM platform, the task simulates a partially observable post-flood search scenario characterized by stochastic dynamic hazards and static obstacles, as shown in Figure~\ref{fig:macsr_scenarios}. The core challenge lies in evaluating the agents' ability to achieve the following objectives under conditions of partial observability:
\begin{itemize}
    \item \textbf{Collaboration:} Agents must collaborate to collect supplies that require multi-agent manipulation.
    \item  \textbf{Safety Constraints:} Agents are required to proactively identify and evade hazardous materials moving randomly within the environment.
    \item  \textbf{Efficient Navigation:} Relying solely on local sensory data, agents must plan energy-efficient, optimal paths through complex, dynamic environments.
\end{itemize}
Fundamentally, MACS serves as a robust experimental platform for the training and evaluation of state-of-the-art multi-agent reinforcement learning (MARL) algorithms. Technically, the environment is formulated with a continuous 2-dimensional action space ($\text{Box}(-1.0, 1.0)$) governing agents' movements, and a high-dimensional observation space constructed from a radial array of sensors that capture the relative distance, orientation, and velocity of surrounding entities. Furthermore, to accommodate diverse experimental requirements, the environment features high configurability. The key parameters and their default settings are detailed in Table \ref{tab:macsr_specs}.

\subsubsection{Baseline}
To validate the efficacy of the proposed benchmark and establish a reference for future research, we evaluate two representative MARL algorithms. These baselines span from fully decentralized approaches to centralized training paradigms.
(1) Independent PPO (IPPO) \cite{de2020independent}: A fully decentralized algorithm where each agent learns an independent policy based solely on its local observation, ignoring the joint state information.
(2) Multi-Agent PPO (MAPPO) \cite{yu2022surprising}: A widely-adopted algorithm following the centralized training with decentralized execution (CTDE) paradigm. It utilizes a centralized value function to exploit global state information during training while maintaining decentralized policies during execution.

We train and evaluate these baselines under the standard environment configuration (default values provided in Table \ref{tab:macsr_specs}).
We employ the mean episodic return per agent ($\bar{R}$) as our primary metric, defined as the total team reward normalized by the agent population $N$. Formally:
\begin{equation}
    \bar{R} = \frac{1}{N} \sum_{i=1}^{N} \sum_{t=1}^{T} r_{t}^{(i)}
    \label{eq:mean_return}
\end{equation}
\noindent where $r_{t}^{(i)}$ denotes the reward received by agent $i$ at time step $t$, and $T$ represents the episode horizon. We additionally report the mean step return, calculated as $\bar{R} / T$, which quantifies the average reward density per time step.

\begin{table}[t]
    \centering
    \caption{Detailed environment specifications and default hyperparameters for the MACS Task.}
    \label{tab:macsr_specs}
    \begin{tabular}{@{}llc@{}}
        \toprule
        \textbf{Parameter} & \textbf{Description} & \textbf{Default} \\
        \midrule
        \texttt{n\_agents} & Number of rescue agents & 5 \\
        \texttt{n\_supplies} & Number of valuable supply items & 10 \\
        \texttt{n\_hazards} & Number of hazard items & 10 \\
        \texttt{n\_coop} & Agents required for successful cooperation & 2 \\
        \texttt{n\_sensors} & Number of sensors per agent & 30 \\
        \texttt{sensor\_range} & Maximum sensing range & 500.0 \\
        \texttt{max\_cycles} & Maximum steps per episode & 500 \\
        \texttt{supply\_reward} & Reward for successfully collecting supplies & 10.0 \\
        \texttt{hazard\_reward} & Penalty for encountering hazards & $-1.0$ \\
        \texttt{encounter\_reward} & Reward for touching a supply without capture & 0.01 \\
        \texttt{thrust\_penalty} & Energy cost multiplier per movement step & $-0.01$ \\
        \texttt{local\_ratio} & Ratio of local rewards to global rewards & 0.9 \\
        \bottomrule
    \end{tabular}
\end{table}

\subsubsection{Experiments and Results}
Table \ref{tab:baseline_results} presents the comparative performance of the baseline algorithms. Empirical results demonstrate that the centralized training approach significantly outperforms the fully decentralized method. MAPPO achieves the highest mean episodic return per agent of 19.24, whereas IPPO yields 14.75. This performance gap indicates that the CTDE paradigm effectively leverages global information to coordinate agents in complex search tasks. Both learning-based methods significantly surpass the Random policy baseline.
\begin{table}[tb]
    \centering
    \caption{Performance comparison of baseline algorithms on the MACS Task. Results are averaged over evaluation episodes.}
    \label{tab:baseline_results}
    \begin{tabular}{@{}lcc@{}}
        \toprule
        \textbf{Method} & \textbf{Mean Step Reward} & \textbf{Mean Episodic Return per Agent} \\
        \midrule
        MAPPO (CTDE) & \textbf{0.0380} & \textbf{19.24} \\
        IPPO (Independent) & 0.0295 & 14.75 \\
        Random & -0.013 & -6.51 \\
        \bottomrule
    \end{tabular}
\end{table}

\subsection{Human-Robot Hybrid Task: Robot Social Navigation}
\subsubsection{Benchmark Overview}

\begin{figure*}[tb]
    \centering
    \includegraphics[width=1.0\linewidth]{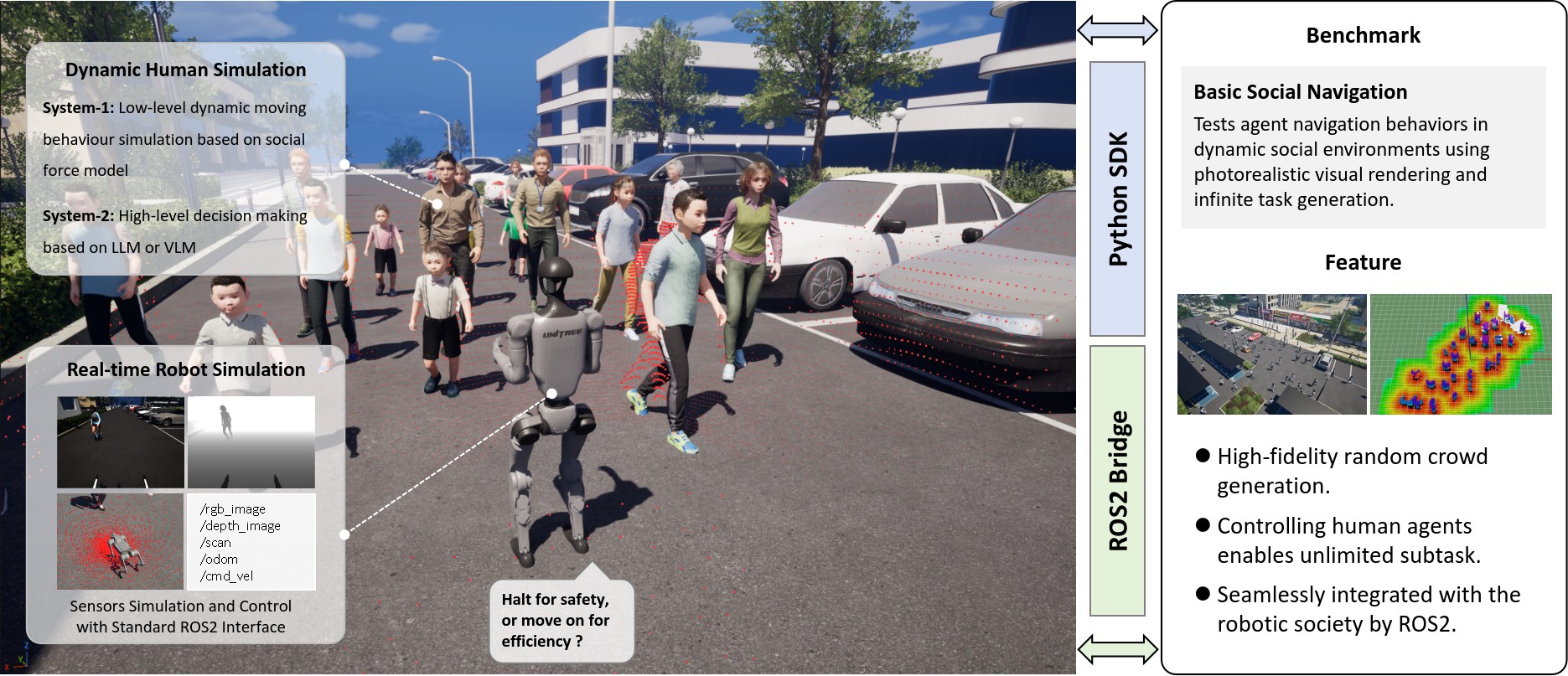}
    \caption{Overview of the robot social navigation task in dynamic social scenarios.
   }
   \label{fig:HRI_bench}
\end{figure*}

Autonomous robots represent the quintessential manifestation of embodied intelligence within the physical world. In the era of human-robot symbiosis, achieving safe and efficient human-centric interaction within large-scale, dynamic, and unstructured environments has emerged as a pivotal scientific challenge in the fields of AGI and robotics.
To advance research in this domain, complex open-world testing environments such as Virtual Community~\cite{zhou2025virtual} and SimWorld~\cite{yesimworld} have been developed. These platforms facilitate the evaluation of intelligent behaviors of embodied agents within complex social scenarios, ranging from autonomously navigating intersections characterized by mixed pedestrian-vehicle traffic to assisting human agents in task execution.
However, Virtual Community is limited to supporting a maximum concurrency of 15-25 human agents and robots within a single scene. Conversely, while SimWorld accommodates a larger population, its navigation relies on predefined discrete waypoints, thereby failing to simulate the continuous crowd dynamics required for rigorously evaluating the performance of realistic autonomous robots.

Leveraging TongSIM's comprehensive asset library and robust control API, we employ the social force model (SFM) to simulate high-fidelity, dynamic crowd behaviors. The system ensures stability for simulations involving over 100 concurrent pedestrians. Virtual robots feature standard ROS2 integration, allowing for seamless interoperability with the broader robotic ecosystem. Furthermore, the system supports human teleoperation of virtual avatars via VR, enabling subjective experience analysis and interaction evaluation.

Within an urban street-level environment, we have designed a fundamental social navigation benchmarking task. This task requires a robot to navigate to a designated target location inside a highly dynamic crowd, characterized by randomly generated agents exhibiting diverse motion primitives. This benchmark aims to evaluate the robot's perception, planning, and social cognitive capabilities within complex social settings, thereby providing a reproducible platform for research in human-robot interaction (HRI). Figure~\ref{fig:HRI_bench} shows an overview of this benchmark.

The robot is tasked with navigating to a designated target location within a specified time limit while traversing a complex environment. The core objective is to generate collision-free trajectories that strictly adhere to proxemic constraints (i.e., maintaining appropriate social distances from pedestrians). This task serves as a benchmark to rigorously evaluate the autonomous agent's perceptual robustness, motion planning efficiency, and social compliance (cognitive understanding of social norms) in dynamic, populated settings.

The robot perceives the environment through a multi-modal sensor suite, including: 1) RGB-D camera, 2) 3D LiDAR, and 3) GPS module.
The decision module outputs continuous control commands via the standard ROS interface.

\subsubsection{Baseline and Metrics}
\textbf{Baseline.} To evaluate performance, we compare the following decision-making approaches.
(1) Human Teleoperation: Expert baseline via manual keyboard control.
(2) Hierarchical planner (DWA)\cite{dwa}: A* global planner \cite{hart1968formal} + Dynamic Window Approach (DWA) local reactive controller.
(3) Hierarchical planner (MPPI)\cite{Williams2015ModelPP}: A* global planner + Model Predictive Path Integral (MPPI) sampling-based local optimizer.

\textbf{Evaluation Metrics.}
We employ a comprehensive set of metrics to quantitatively evaluate performance. The evaluation module operates as a background process with a sampling frequency of 2 Hz.
\begin{itemize}
    \item \textbf{Efficiency (EFF):} Measures the time efficiency of the navigation task based on the \textbf{actual completion time ($T_{actual}$)}, normalized by the theoretical minimum time ($T_{min}$) and the maximum allotted time ($T_{max}$). It is calculated as:
    \begin{equation}
        \text{EFF} = 1 - \frac{T_{actual} - T_{min}}{T_{max} - T_{min}}
    \end{equation}
    \item \textbf{Success Rate (SRT):} The ratio of successfully completed episodes to the total number of experimental trials.
    \item \textbf{Safety (SAF):} A discrete scoring metric reflecting collision frequency. We assign a score of 1.0 for zero collisions, 0.5 for 1--3 collisions, and 0.0 for more than 3 collisions.
    \item \textbf{Social Norm Compliance (SNC):}
    quantifies the percentage of task time during which the robot intrudes into a human’s personal space, distinguishing Type-1 intrusions (distance $d < 0.45\,\text{m}$) and Type-2 intrusions (distance $0.45\,\text{m} \le d \le 1.2\,\text{m}$).
    \item \textbf{Total Score:} A weighted aggregate score quantifying overall performance:
    \begin{equation}
        \text{Total} = 100 \times (0.2 \cdot \text{EFF} + 0.2 \cdot \text{SRT} + 0.3 \cdot \text{SAF} + 0.3 \cdot \text{SNC})
    \end{equation}
\end{itemize}
\subsubsection{Experiments and Results}
\begin{table}[htbp]
    \centering
    \caption{Performance comparison of baseline methods in the crowded intersection crossing task. 
    The test environment consists of a circular area ($r=20\,\text{m}$) populated with 30 randomly moving pedestrians, with a minimum start-to-goal distance of 40\,\text{m}.}
    \label{tab:intersection_performance}
    \begin{tabular}{l c c c c c c}
        \toprule
        \textbf{Baseline} & \textbf{Robot} & \textbf{EFF} & \textbf{SRT} & \textbf{SAF} & \textbf{SNC} & \textbf{Total} \\
        \midrule
        Human Teleoperation & Unitree Go2 & \textbf{0.89} & \textbf{1.0} & \textbf{0.95} & \textbf{0.88} & \textbf{92.7} \\
        \midrule
        $A^*$ + DWA & Unitree Go2 & 0.42 & 0.1 & 0 & 0 & 10.4 \\
        $A^*$ + MPPI & Unitree Go2 & 0.73 & 0.6 & 0.25 & 0.31 & 43.1 \\
        \bottomrule
    \end{tabular}
\end{table}
As presented in Table \ref{tab:intersection_performance}, there is a marked disparity in performance across the evaluated decision-making baselines within the fundamental social navigation task. Human Teleoperation achieved superior performance across all four metrics (Total Score = 92.7), underscoring the human capacity to leverage contextual reasoning and social cognition for high-quality navigation amidst complex crowd dynamics. In stark contrast, global $A^*$ with local DWA planner exhibited instability in highly dynamic scenarios, recording an exceptionally low success rate (SRT = 0.1). The method suffered from frequent collisions and severe spatial intrusions, resulting in zero scores for both safety (SAF) and social norm compliance (SNC), yielding a total score of only 10.4. This highlights the inadequacy of local planners relying solely on geometric constraints when confronting the stochastic nature of multi-agent behaviors. While global $A^*$ with local MPPI planner demonstrated improvements in efficiency and success rates through sampling-based optimization (achieving moderate performance in safety and social compliance, and total score = 43.1) it still falls significantly short of human benchmarks. This performance gap indicates that traditional planning frameworks, lacking social behavioral understanding, struggle to adapt to dense crowd interactions.

The quantitative experimental results validate a core conclusion: in complex, dynamic social environments, the absence of social contextual reasoning and normative cognition causes traditional autonomous planning methods to systematically lag behind human performance in terms of safety, success rate, and social compliance. Embodied agents relying exclusively on geometric obstacle avoidance or local optimization cannot achieve reliable and socially normative interactive behaviors. This critical limitation emphasizes the necessity of integrating models endowed with social cognitive capabilities.

\subsection{Primary Composite Tasks: Household Benchmarking Test}
\subsubsection{Benchmark Overview}
Recent advancements in multimodal large language models (MLLMs) have significantly enhanced AI capabilities in perception and linguistic comprehension. However, existing benchmarks (e.g., ImageNet \cite{deng2009imagenet}, COCO \cite{lin2014microsoft}, VQA \cite{antol2015vqa}) predominantly focus on isolated, domain-specific tasks. These benchmarks suffer from limitations regarding hyperspecialization and susceptibility to overfitting, making them insufficient for comprehensively gauging a model's capabilities in open-ended, dynamic real-world environments—competencies essential for the progression towards AGI. Although multi-task platforms like MMBench \cite{liu2024mmbench} offer multidimensional testing, they generally lack embodied interaction capabilities, failing to evaluate model reasoning and action execution within physical and social contexts.

\begin{figure*}[tb]
    \centering
    \includegraphics[width=1.0\linewidth]{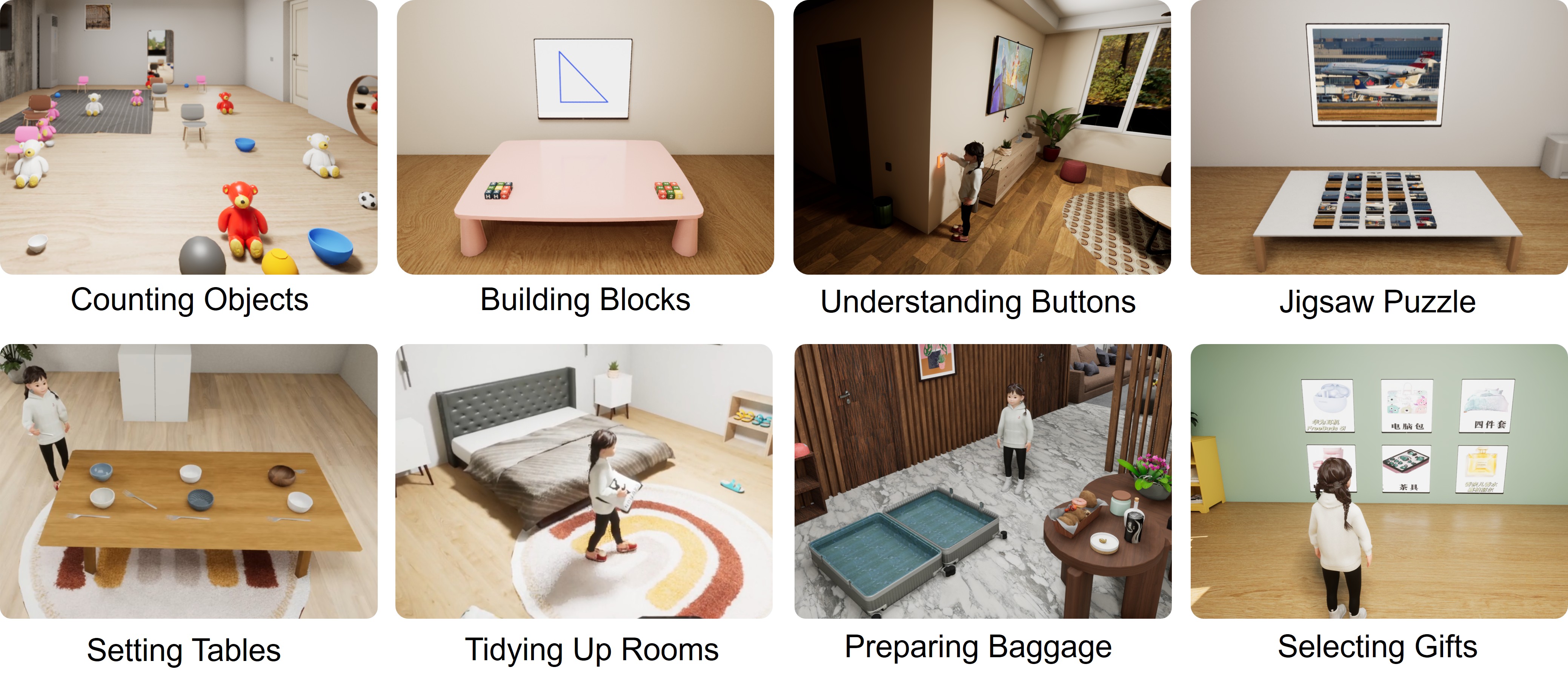}
    \caption{Overview of the primary family composite tasks benchmark.  }
    \label{fig:benchmark_tasks}
\end{figure*}

To address this critical limitation, we propose a novel evaluation paradigm: an embodied composite task benchmark situated in everyday household environments and drawing inspiration from early childhood developmental psychology, as shown in Figure \ref{fig:benchmark_tasks}. The core objective of this benchmark is to evaluate how MLLM agents integrate perception, reasoning, and action to accomplish composite tasks requiring synergistic capabilities, thereby providing a more realistic assessment of their general intelligence.

\subsubsection{Test Procedure}
Focusing on composite task scenarios frequent in household settings, we designed an evaluation set comprising eight task types, categorized into three core domains:
\begin{itemize}
    \item \textbf{Object Understanding:} Including object counting and gift selection.
    \item \textbf{Spatial Intelligence:} Including block construction, puzzle solving, and understanding button functionality.
    \item \textbf{Social Activity:} Including table setting, room organization, and luggage preparation.
\end{itemize}
We construct high-fidelity 3D virtual home environments within the TongSIM simulation platform to serve as a testbed for model evaluation. Leveraging prompt engineering, we encapsulate the target models into agents capable of executing tasks within these embodied interactive settings. Consequently, we establish a standardized perception-reasoning-action loop framework for the MLLMs under assessment:
\begin{itemize}
    \item \textbf{Perception:} At each time step, the agent receives observational data from the environment, comprising multi-view RGB images and JSON scene descriptions containing object attributes (e.g., name, color, position).
    \item \textbf{Reasoning and Decision-Making:} Based on observational data and natural language task goals, the MLLM employs a ReAct-style reasoning process to output a natural language reasoning trace and an executable sequence of API calls.
    \item \textbf{Action:} The system executes the API call sequence (e.g., MoveToObject, PickUp), which encompasses both atomic and high-level actions, thereby driving the virtual character to interact with the environment.
    \item \textbf{Loop:} This process iterates until the task is completed or a timeout occurs.
\end{itemize}
To directly assess the intrinsic capabilities of the models, we did not perform retraining or introduce external augmentation modules; instead, we encapsulated the MLLMs as embodied agents solely through carefully designed prompts.

\subsubsection{Experiments and Results}
We conducted a systematic evaluation of 17 proprietary and open-source MLLMs (including the GPT, Gemini, Claude, and Llama series). The primary results are as follows:

\begin{table*}[b]
\centering
\caption{Model performance comparison on the primary family composite tasks benchmark.}
\label{tab:model_comparison_primary}
\resizebox{\textwidth}{!}{%
\begin{tabular}{l*{10}{c}} 
\toprule
\multirow{3}{*}{\textbf{Model}} & 
\multicolumn{2}{c}{\textbf{Object Understanding}} & 
\multicolumn{3}{c}{\textbf{Spatial Intelligence}} & 
\multicolumn{3}{c}{\textbf{Social Activity}} & 
\multirow{3}{*}{\textbf{Mean}} \\
\cmidrule(lr){2-3} \cmidrule(lr){4-6} \cmidrule(lr){7-9}
& \shortstack{\textbf{Counting} \\ \textbf{Objects}} & \shortstack{\textbf{Selecting} \\ \textbf{Gifts}} & \shortstack{\textbf{Building} \\ \textbf{Blocks}} & \shortstack{\textbf{Jigsaw} \\ \textbf{Puzzle}} & \shortstack{\textbf{Understanding} \\ \textbf{Buttons}} & \shortstack{\textbf{Setting} \\ \textbf{Tables}} & 
\shortstack{\textbf{Tidying Up} \\ \textbf{Rooms}} & \shortstack{\textbf{Preparing} \\ \textbf{Baggage}} & \\
\midrule

Gemini-2.5-Pro	        &48.00 	&68.06 	&10.00 	&5.05 	&3.33 	&26.68 	&22.77 	&\textbf{12.38} 	&\textbf{24.53} \\
Gemini-2.5-Flash	    &42.00 	&68.20 	&5.50 	&5.30 	&3.33 	&25.83 	&\textbf{23.22} 	&11.05 	&23.05 \\
o3	                    &\textbf{54.00}  &65.92 	&10.00 	&6.40 	&3.33 	&14.31 	&18.77 	&10.30 	&22.88 \\
GPT-5	                &36.00 	&\textbf{69.06} 	&3.75 	&6.03 	&3.33 	&\textbf{28.68} 	&16.00 	&9.50 	&21.54 \\
Claude-3.7-Sonnet	    &46.00 	&59.74 	&8.88 	&6.28 	&0.00 	&23.76 	&16.12 	&3.40 	&20.52 \\
Claude-4-Sonnet	        &44.00 	&65.44 	&8.75 	&6.03 	&0.00 	&19.81 	&14.85 	&5.18 	&20.51 \\
Doubao-1.5-vision-pro	&36.00 	&56.70 	&\textbf{12.50} 	&6.88 	&\textbf{5.00} 	&24.17 	&4.22 	&7.75 	&19.15 \\
Claude-3.5-Sonnet	    &42.00 	&64.24 	&8.00 	&4.50 	&0.00 	&12.33 	&13.47 	&6.00 	&18.82 \\
Grok 3	                &52.00 	&50.24 	&9.88 	&\textbf{8.00} 	&0.00 	&8.37 	&17.20 	&3.35 	&18.63 \\
o4-mini	                &38.00 	&66.40 	&0.00 	&6.62 	&3.33 	&13.67 	&1.52 	&4.23 	&16.72 \\
GPT-4o 	                &32.00 	&46.30 	&10.00 	&7.08 	&3.33 	&18.42 	&6.30 	&8.18 	&16.45 \\
GPT-4o-mini	            &34.00 	&54.96 	&5.75 	&7.05 	&1.67 	&19.07 	&1.50 	&3.75 	&15.97 \\
Qwen-VL-max	            &44.00 	&48.14 	&0.00 	&7.55 	&0.00 	&15.96 	&1.25 	&0.00 	&14.61 \\
Llama-4-Maverick	    &36.00 	&50.30 	&3.75 	&6.58 	&0.00 	&15.75 	&0.42 	&3.00 	&14.48 \\
Llama-4-Scout	        &28.00 	&42.56 	&6.25 	&5.72 	&0.00 	&8.94 	&1.38 	&2.00 	&11.86 \\
Qwen-VL-plus	        &6.00 	&34.16 	&0.00 	&7.65 	&0.00 	&16.67 	&1.82 	&0.38 	&8.33 \\
Llama-3.2	            &6.00 	&6.80 	&0.00 	&4.60 	&0.00 	&4.67 	&0.42 	&0.00 	&2.81 \\
\bottomrule
\end{tabular}%
}
\end{table*}
\begin{itemize}
    \item \textbf{Overall Suboptimal Performance:} Average scores across all models on composite tasks were universally low. The top-performing model, Gemini-2.5-Pro, achieved an average score of merely 24.53/100. This indicates a substantial gap between state-of-the-art MLLMs and the requirements for handling real-world embodied tasks.
    \item \textbf{Proprietary vs. Open-Source Models:} While proprietary models generally outperformed open-source counterparts, the margin was not overwhelming. The best-performing open-source model was Llama-4-Maverick, with a score of 14.48.
    \item \textbf{Domain-Specific Analysis:}
    (1) Object Understanding proved to be the relative strength of the models; for instance, GPT-5 achieved a score of 69.06 in the ``Gift Selection'' task. This suggests that current models possess certain advantages in perception and basic classification tasks.
    (2) Spatial Intelligence emerged as the most challenging domain, with models scoring extremely low on tasks such as block construction and puzzle solving. This identifies spatial reasoning, physical understanding, and manipulation planning as the weakest links in current MLLMs.
    (3) Social Activity performance was inconsistent, with different models leading in different tasks; however, even the highest scores hovered in the 20-30 range. This reflects difficulties in comprehending complex social contexts and executing goal-oriented actions.
\end{itemize}
Experimental results, shown in Table~\ref{tab:model_comparison_primary}, demonstrate that while current MLLMs have made significant strides in perception and language tasks, their architectures, training mechanisms, and multimodal fusion strategies remain insufficient to support the composite capabilities required for embodied tasks. Future research must prioritize the in-depth development of embodied reasoning, spatial intelligence, and social understanding. The benchmark proposed in this work serves as a vital evaluation tool and provides directional guidance for advancing MLLMs toward general embodied intelligence.

\subsection{Advanced Composite Tasks: Spatially Situated Social Intelligence Test}
\begin{figure}[t]
  \centering
  \includegraphics[width=1.0\linewidth]{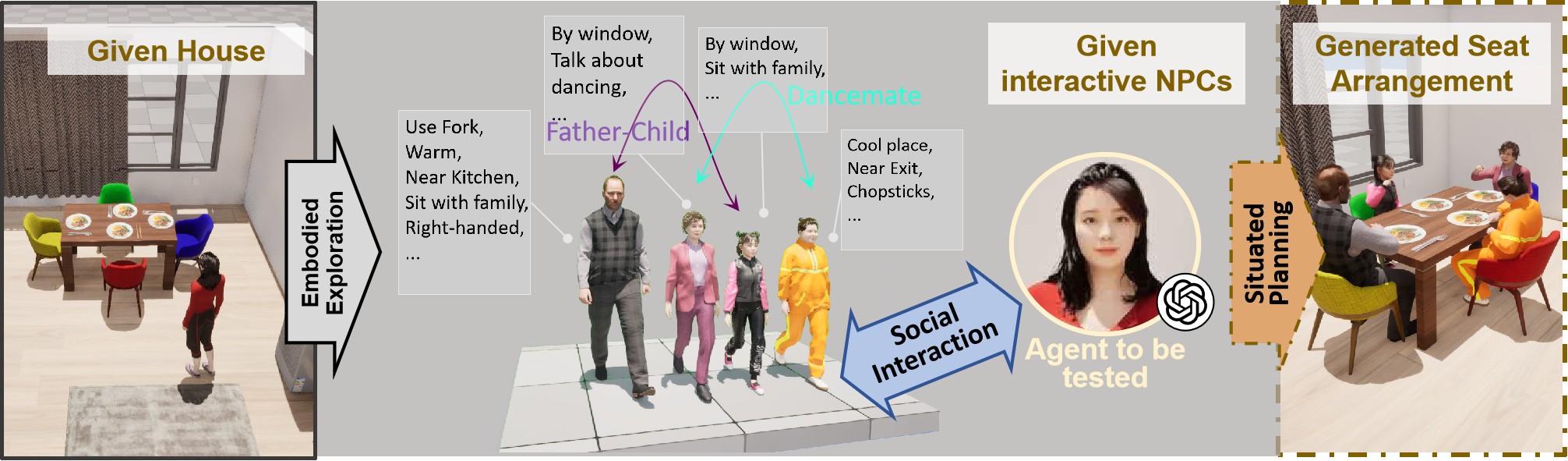} 
  \caption{A typical seat arrangement task involves a given room layout and several NPCs. The agent under test (T-Agent) needs to interact with the NPCs and explore the room to devise a seating arrangement that satisfies everyone.}
  \label{fig:s3it}
\end{figure}

\subsubsection{Benchmark Overview}
The integration of embodied agents into human environments demands embodied social intelligence: reasoning over both social norms and physical constraints. However, existing evaluations fail to address this integration, as they are limited to either disembodied social reasoning (e.g., in text) or socially-agnostic physical tasks. Both approaches fail to assess an agent's ability to integrate and trade off both physical and social constraints within a realistic, embodied context.

To address this challenge, we introduce spatially situated social intelligence test (S$^3$IT), a benchmark specifically designed to evaluate embodied social intelligence. As shown in Figure~\ref{fig:s3it}, it is centered on a novel and challenging seat-ordering task, requiring an agent to arrange seating in a 3D environment for a group of LLM-driven NPCs with diverse identities, preferences, and intricate interpersonal relationships. Our procedurally extensible framework generates a vast and diverse scenario space with controllable difficulty, compelling the agent to acquire preferences through active dialogue, perceive the environment via autonomous exploration, and perform multi-objective optimization within a complex constraint network.

In this benchmark, we constructed 5 scene environments with dynamic layout characteristics
, where key facilities (e.g., chairs, furniture layouts) support parametric configuration. Simultaneously, we established a virtual community comprising 59 NPCs with independent background settings, and defined their internal complex family and social relationship networks. Based on this, we constructed 7,000 problems. Each problem includes one room and several NPCs; while basic information remains invariant, we additionally configured specific hobbies, seating preferences, and conflicts with others for each NPC in every problem. Furthermore, we assigned 3-Likert point intensities (weights) to these preferences and conflicts to enrich the problems.

\subsubsection{Baselines and Test Procedure}
Within this benchmark, we propose a testing framework designed to integrate LLMs. To enable the agents under test to effectively comprehend and execute the seating arrangement tasks, we have systematically designed and standardized the inputs, outputs, and prompt structures. The pipeline comprises three phases for testing the agent that is named as ``T-Agent''. In Phase I (NPC Preference Extraction and Summarization), the T-Agent constructs detailed preference profiles for each NPC. Then, in Phase II (Environmental Cognition ), the T-Agent needs to construct a structured representation of the 3D environment through comprehensive exploration. Finally, in Phase III (Multi-Constraint Decision-Making), the T-Agent integrates information from the preceding phases to generate, reflect, and iteratively refine seating solutions. To enable scalable evaluation, we introduce an automated scoring framework. This framework evaluates the preference satisfaction of agent-generated seating plans, thereby providing an objective, quantitative metric for each solution. Furthermore, to analyze the T-Agent's capability in prioritizing based on weights, we statistically tracked its preference satisfaction based on their strengths. We define the prioritization gap (PG) as: 
\begin{equation}
    \text{PG} = S_{\text{high}} - S_{\text{low}}
    \label{eq:pg_metric}
\end{equation}
where $S_{\text{high}}$ denotes the satisfying rate of high-weighted preferences, and $S_{\text{low}}$ denotes the satisfying rate of low-weighted preferences.

\subsubsection{Experiments and Results}

\begin{table}[tb]
\centering
\caption{Model performances on the S$^3$IT benchmark's test set. Best results are in bold. PG stands for the prioritization gap.}
\label{tab:results_compact}
\setlength{\tabcolsep}{8pt}
\begin{tabular}{lccccc}
\toprule
\textbf{Model} & \textbf{Embodied} & \textbf{Social}  & \textbf{Conflict} &\textbf{PG} & \textbf{Average} \\
\midrule
Gemini-2.5-pro & \textbf{40.6} & 56.2 & 85.7 & 8.8& \textbf{47.8} \\
o3 & 32.9 & 53.8 & 89.0 &12.7 & 43.1 \\
GPT-5 & 29.0 & \textbf{56.9} & 86.1 &\textbf{15.4} & 42.7 \\
o4-mini & 29.0 & 54.5 & \textbf{89.5} &6.8 & 41.4 \\
GPT-4.1 & 23.3 & 43.2 & 55.4 &3.8 & 29.3 \\
Doubao-1.5 & 24.6 & 43.0 & 62.5 &3.8 & 28.3 \\
GPT-4o & 24.6 & 43.0 & 51.7 &3.3 & 28.2 \\
GPT-4.1-mini & 22.8 & 39.3 & 42.5 &3.7 & 26.8 \\
Claude-4.5 & 19.1 & 37.6 & 46.0 &4.8 & 23.1 \\
GPT-4o-mini & 16.7 & 34.7 & 45.8 &1.6 & 19.3 \\
\bottomrule
\end{tabular}
\end{table}

As shown in Table \ref{tab:results_compact}, Gemini-2.5-pro emerged as the SOTA model (47.8) and was uniquely capable of exceeding 40 on the ``Embodied Preference'' dimension. These models exhibit a consistent trend that scores are lowest in the embodied dimension, followed by the social dimension. In contrast, performance in the conflict dimension is generally strong across the board, with some models even demonstrating a good capacity for conflict resolution. These findings validate the hypothesis that spatial intelligence is the cornerstone of effective embodied social reasoning. All models demonstrated positive PG scores, indicating that they have a fundamental ability to distinguish preference strengths. Higher-priority preferences are fulfilled with greater precedence. Among the evaluated models, GPT-5, with an updated adaptive reasoning architecture, excels at considering preferences of varying intensities.

\section{Discussion}

\subsection{Various benchmarks}
TongSIM is established not merely as a simulation platform, but as a holistic proving ground aimed at catalyzing agent evolution. Underpinned by a versatile suite of benchmarks, our multi-dimensional architectural philosophy enables a full-spectrum diagnosis and validation of agent proficiency.

First, a distinguishing feature of our platform is the hierarchical evaluation of agent capabilities. Our benchmarks span a comprehensive spectrum, ranging from low-level perception and locomotion to high-level complex tasks and social intelligence.
At the foundational level, we assess atomic single-agent capabilities, including locomotion, steering, and basic visual perception, which are incorporated into all our proposed benchmarks. Building upon this, single-agent tasks involve higher-level assessments such as object attribute recognition, spatial memory construction, and semantic navigation. For instance, requiring an agent to explore an unknown room to locate a ``paper ball" tests its semantic understanding of the environment.
At the highest level, the platform emphasizes sociality and collaborative intelligence. This encompasses high-dimensional challenges ranging from multi-agent collaboration (e.g., the MACS task) to complex scenarios (e.g., navigation in dense crowds and execution of complex household tasks). These benchmarks aim to advance the development of agents towards AGI equipped with social attributes.

Second, high-fidelity simulation scenarios are pivotal to bridging the Sim-to-Real gap. In contrast to traditional grid worlds or symbolic, low-fidelity environments, all tasks within our platform are situated in photorealistic environments constructed upon advanced physics engines and ray-tracing technologies.
TongSIM scenarios encompass complex illumination variations, specular reflections, shadow occlusions, and rich textural details, thereby compelling the agents' visual systems to demonstrate robustness against real-world sensory noise.
Furthermore, TongSIM features rich physical simulation capabilities, enabling the modeling of rigid body dynamics, fluid interactions, and deformable object deformations (e.g., cloth and liquids). Under such constraints, the planning capabilities of agents can be evaluated within scenarios that closely approximate reality.
For instance, when grasping a cup filled with water, the agent must account for shifts in the center of gravity and friction. This training regime ensures that model parameters, upon transfer to physical robotic hardware, can directly adapt to real-world physical constraints, significantly mitigating the costs associated with fine-tuning. Although this paper does not go into detail about this feature, the experimental development work has been basically completed, which will be introduced in subsequent work.

We prioritize agent generalizability and are dedicated to the pursuit of AGI; accordingly, TongSIM is engineered to support massive, large-scale tasks to facilitate this objective.
To prevent agents from overfitting to specific spatial configurations, the platform is designed to enable large-scale scene generation. Leveraging procedural generation, we can synthesize thousands of indoor environments featuring diverse layouts and stylistic variations.
This diversity ensures that agents acquire generalized interaction logic (e.g., understanding the affordance that a door requires pushing or pulling) rather than relying on the rote memorization of map coordinates. An extensive suite of additional benchmarks will be progressively released to the research community.

\subsection{Potential Applications}
In light of our ultimate objective to achieve AGI, the inception and architectural design of this platform are intrinsically aligned with this vision. Rather than merely supplying datasets, we provide a holistic ecosystem designed to incubate high-level intelligence.

TongSIM offers users an exceptional degree of freedom and extensibility, thereby broadening its multidisciplinary application horizons:
\begin{itemize}
    \item \textbf{Planning in Symbolic Space:} Researchers can leverage high-level APIs to focus on task decomposition, causal reasoning, and the application of knowledge graphs, without being encumbered by low-level control dynamics.
    \item \textbf{End-to-End Embodied Control:} Researchers in reinforcement learning and imitation learning can utilize the comprehensive interfaces and rich environmental data (e.g., RGB-D images, semantic object labels) to train end-to-end neural networks.
    \item \textbf{Verification of Hybrid Architectures:} The platform facilitates the validation of ``LLM-Brain + Controller-Cerebellum'' hybrid architectures. For instance, users can employ LLMs for high-level strategic planning while invoking low-level motion primitives to execute specific operations.
\end{itemize}
To transcend the limitations of traditional platforms restricted to fixed tasks, we provide standardized Python/C++ APIs, diverse virtual avatars, and extensive functional capabilities. These tools enable users to rapidly construct custom tasks:
\begin{itemize}
    \item \textbf{Domestic Service Scenarios:} Users can define a ``room tidying'' task, requiring the agent to identify debris on the floor, classify items, and place them into storage containers.
    \item \textbf{Operations in Extreme Environments:} Simulating post-fire ruin scenarios to train agents in Search and Rescue and hazardous material removal, testing their robustness within chaotic environments.
    \item \textbf{Human-Robot Interaction:} Configuring NPCs with social attributes to train agents in comprehending human gestures and gaze intent, as well as engaging in natural language dialogue.
\end{itemize}
These tasks are not confined to the platform's preset benchmarks; researchers can rapidly construct experimental environments tailored to their specific scientific hypotheses, thereby significantly enhancing iteration efficiency.

\subsection{Future Works}
Currently, the platform is undergoing iterative development. We are committed to progressively refining both indoor and outdoor environments, expanding the benchmark suite, and releasing these contributions to the open-source community.
Regarding environmental expansion, while our current focus lies within complex indoor domestic environments, we are actively constructing an expansive world model. Future iterations will extend into outdoor scenarios, enriching supported interactions and semantic annotations. This includes the simulation of urban street interactions, crowd and traffic network dynamics, unstructured natural terrains (e.g., forests, mountains), and specialized outdoor zones (e.g., industrial complexes). These developments are poised to support research in autonomous driving, logistics distribution, and field exploration robotics. Furthermore, we plan to enhance the temporal and environmental dynamics of the simulator. Future scenes will transcend static snapshots by incorporating diurnal cycles, meteorological variations (e.g., rain, fog, snow), and dynamic obstacles (e.g., pedestrians, vehicles), thereby approximating the stochasticity and complexity of the real world.

The future of embodied AI is inextricably linked to human cognitive guidance and cross-reality interaction. The emerging paradigm of symmetrical reality \cite{zhang2024emergence,zhang2019symmetrical}, utilizing human-in-the-loop methodologies across physical and virtual worlds, grants agents direct exposure to data distributions of hybrid environments, fostering robust training for extensible tasks.

Through these endeavors, we envision this platform not merely as a testing tool, but as a stepping stone toward a future of human-machine symbiosis, contributing substantive momentum to the realization of AGI endowed with comprehensive perception, reasoning, and actuation capabilities.

\section{Conclusions}

This paper formally introduces TongSIM, a universal training and evaluation platform for embodied AI designed to bridge the simulation-to-reality gap through high-fidelity simulation technologies. TongSIM establishes an expansive world featuring 115 meticulously crafted interactive indoor scenes and large-scale continuous urban environments, and provides highly challenging multimodal sensory inputs to agents. Leveraging high-fidelity physics simulation, native parallel training support, and diverse agent driving modalities, TongSIM establishes a task spectrum that surpasses existing platforms. 

We design and implement five benchmark categories within the 3D interactive environments, covering single-agent autonomous navigation, multi-agent cooperation, social navigation in human-robot hybrid environments, and both basic and advanced household service tasks. Extensive empirical evaluations reveal that while agents driven by reinforcement learning and MLLMs excel in specific tasks, they exhibit significant deficiencies in long-horizon planning, complex spatial reasoning, and adherence to social norms.

To foster community development, we release TongSIM as an open-source platform. We envision this infrastructure serving as a catalyst for general embodied intelligence research, empowering both academia and industry to cultivate next-generation agents capable of robust perception, deep reasoning, and efficient cooperation within open and dynamic environments.




\end{document}